\documentclass[final]{cvpr}

\usepackage{times}
\usepackage{epsfig}
\usepackage{graphicx}
\usepackage{amsmath}
\usepackage{amssymb}

\usepackage[utf8]{inputenc} 
\usepackage{pifont}
\usepackage{amsfonts,amssymb,nicefrac,array,mathtools}     
\usepackage{microtype} 
\usepackage{graphicx,epsfig,subcaption}
\usepackage{multirow,multicol,booktabs}
\usepackage[flushleft]{threeparttable}
\usepackage{color}
\usepackage{comment}
\usepackage{algorithmic}
\usepackage[ruled,vlined]{algorithm2e}
\usepackage{mwe}
\usepackage{adjustbox}
\usepackage{enumitem}
\usepackage[table]{xcolor}

\usepackage[belowskip=-8pt,aboveskip=13pt]{caption}
\newcommand{\cmark}{\text{\ding{51}}}%
\newcommand{\xmark}{\text{\ding{55}}}%

\newcommand{\tablestyle}[2]{\setlength{\tabcolsep}{#1}\renewcommand{\arraystretch}{#2}\centering\footnotesize}

\newcommand{\app}{\raise.17ex\hbox{$\scriptstyle\sim$}}

\newlength\savewidth\newcommand\shline{\noalign{\global\savewidth\arrayrulewidth
  \global\arrayrulewidth 1pt}\hline\noalign{\global\arrayrulewidth\savewidth}}
  
\makeatletter\renewcommand\paragraph{\@startsection{paragraph}{4}{\z@}
  {.5em \@plus1ex \@minus.2ex}{-.5em}{\normalfont\normalsize\bfseries}}\makeatother

\def\fig#1{Fig.~\ref{fig:#1}}
\def\imw#1#2{\includegraphics[clip,width=#2\linewidth]{#1}}

\newcommand{\tb}[3]{\setlength{\tabcolsep}{#2mm}\begin{tabular}{#1}#3\end{tabular}}

\definecolor{ForestGreen}{rgb}{0.13, 0.55, 0.13}
\definecolor{Green}{rgb}{0.0, 0.5, 0.0}
\definecolor{green(munsell)}{rgb}{0.0, 0.66, 0.47}
\definecolor{green(ryb)}{rgb}{0.4, 0.69, 0.2}
\definecolor{green(pigment)}{rgb}{0.0, 0.65, 0.31}

\usepackage[pagebackref=true,breaklinks=true,colorlinks,bookmarks=false]{hyperref}

\def\cvprPaperID{4591} 
\def\confYear{CVPR 2021}

\begin{document}

\title{Unsupervised Feature Learning by Cross-Level Instance-Group Discrimination}

\pagenumbering{gobble}

\author{Xudong Wang\\
UC Berkeley / ICSI \\
{\tt\small xdwang@eecs.berkeley.edu}
\and
Ziwei Liu\\
S-Lab, NTU \\
{\tt\small ziwei.liu@ntu.edu.sg}
\and
Stella X. Yu\\
UC Berkeley / ICSI \\
{\tt\small stellayu@berkeley.edu}
}

\maketitle

\begin{abstract}
Unsupervised feature learning has made great strides with contrastive learning based on instance discrimination and invariant mapping,  as benchmarked on curated class-balanced datasets.  However, natural data could be highly correlated and long-tail distributed.
Natural between-instance similarity conflicts with the presumed instance distinction, causing unstable training and poor performance.

Our idea is to discover and integrate between-instance similarity into contrastive learning, not directly by instance grouping, but by cross-level discrimination (CLD) between instances and local instance groups.  
While invariant mapping of each instance is imposed by attraction within its augmented views, between-instance similarity could emerge from common repulsion against instance groups.  

Our batch-wise and cross-view comparisons also greatly improve the positive/negative sample ratio of contrastive learning and achieve better invariant mapping.  To effect both grouping and discrimination objectives, we impose them on features separately derived from a shared representation.  In addition, we propose normalized projection heads and unsupervised hyper-parameter tuning for the first time.

Our extensive experimentation demonstrates that CLD is a lean and powerful add-on to existing methods  such as  NPID, MoCo,  InfoMin, and BYOL on highly correlated, long-tail, or balanced datasets.  It not only achieves new state-of-the-art on self-supervision, semi-supervision, and transfer learning benchmarks, but also beats MoCo v2 and SimCLR on every reported performance attained with a much larger compute.  
CLD effectively brings unsupervised learning closer to natural data and real-world applications. Our code is publicly available at: \href{https://github.com/frank-xwang/CLD-UnsupervisedLearning}{https://github.com/frank-xwang/CLD-UnsupervisedLearning}.
\end{abstract}

\def\figteaser#1{
\begin{figure}[#1]
\centering
\tb{@{}ccc@{l}}{0}{
\imw{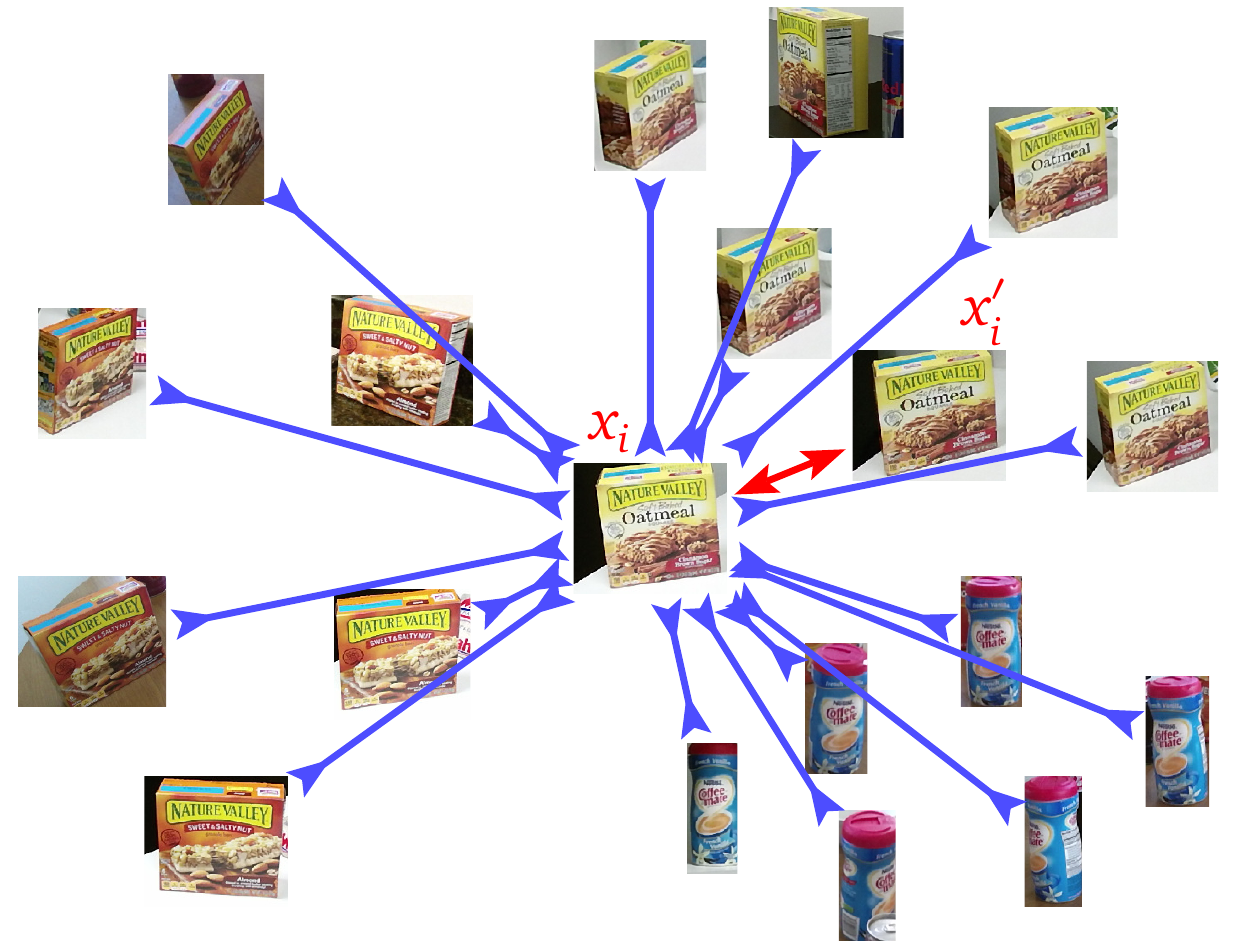}{0.49}&
\imw{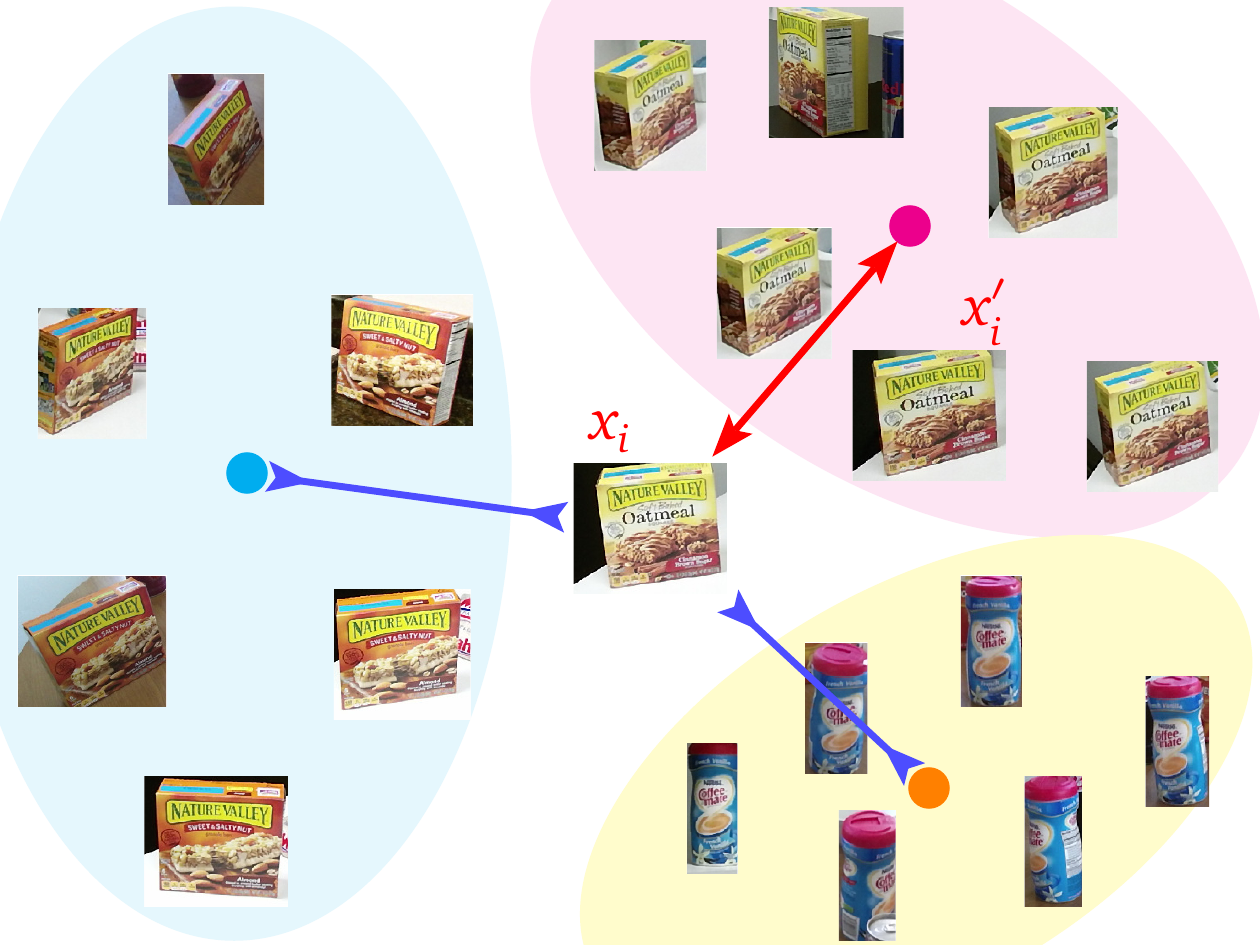}{0.49}\\
{\bf a)} {\small instance discrimination}&
{\bf b)} {\small instance-group discrimination}\\
}\\
\imw{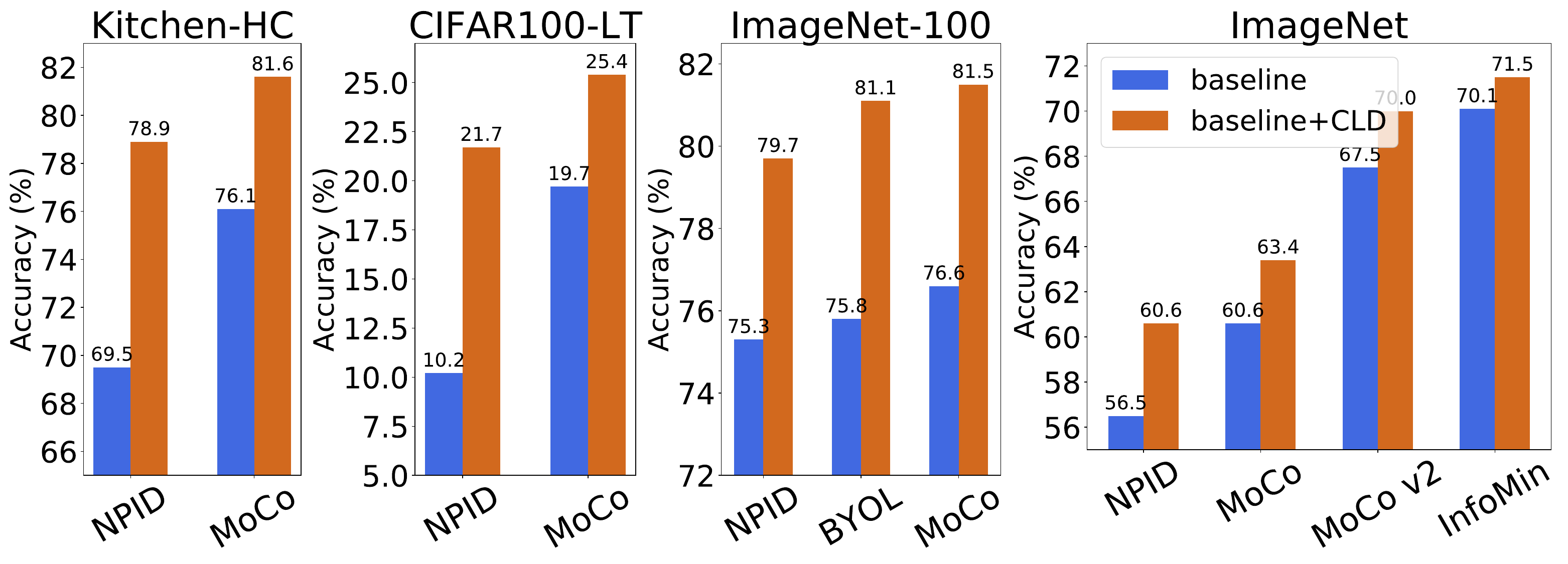}{1.0}\\
{\bf c)} {\small Our cross-level discrimination boosts existing SOTA's}\\\vspace{-6pt}
\caption{Our unsupervised feature learning discovers similar instances and integrates grouping into instance-level discrimination, outperforming the state-of-the-art (SOTA)  classifiers on highly correlated, long-tail, or balanced datasets.
 {\bf a)} Instance discrimination presumes all instances distinctive: Instance $x_i$ \textcolor{red}{attracts ($\leftrightarrow$)} its augmented version $x_i'$ and \textcolor{blue}{repels (${\tiny>} \!\!\!-\!\!\!{\tiny<}$)
 } all other instances including those highly similar ones.  
 {\bf b)} We propose cross-level discrimination (CLD) between instance $x_i$ and local groups of alternative views $\{ x_j' \}$.
 $x_i$ \textcolor{red}{attracts ($\leftrightarrow$)} the group centroid that $x_i'$ belongs to and \textcolor{blue}{repels (${\tiny>} \!\!\!-\!\!\!{\tiny<}$)} other group centroids.  Visually similar instances tend to attract/repel the same group centroids and are thus mapped closer.
 {\bf c)} Our CLD can be added to existing methods such as NPID \cite{wu2018unsupervised}, MoCo \cite{he2020momentum}, MoCo v2 \cite{chen2020improved}, InfoMin \cite{tian2020makes} and BYOL \cite{grill2020bootstrap}.  It consistently provides a significant performance boost on highly correlated (HC), long-tail (LT), and standard balanced ImageNet  datasets. 
}
\label{fig:teaser}
\end{figure}
}





\section{Introduction}

Representation learning aims to extract latent or semantic information from raw data.  Typically, a model is first trained on a large-scale annotated dataset ~\cite{krizhevsky2012imagenet}
and then tuned on a small-scale dataset for a downstream task ~\cite{he2017mask}.
As the model gets bigger and deeper ~\cite{he2016deep, huang2017densely}, more annotated data are needed;
supervised pre-training is no longer viable.

Self-supervised learning~\cite{doersch2015unsupervised, pathak2016context, zhang2016colorful, noroozi2016unsupervised, donahue2017adversarial,misra2019self} gets around labeling with
a pre-text task which does not require annotations and yet would be better accomplished with semantics.  For example,
to predict the color of an object from its grayscale image does not require labeling; however, doing it well would require a sense of what the object is.  The biggest drawback is that pre-text tasks are domain-specific and hand-designed, and they are not directly related to downstream semantic classification.

Unsupervised contrastive learning has emerged as a direct winning alternative
~\cite{wu2018unsupervised,zhuang2019local,ye2019unsupervised,chen2020simple,he2020momentum}.  The training objective and the downstream classification are aligned on discrimination, albeit at different levels of granularities: training is to discriminate known individual instances, whereas testing is to discriminate unknown  groups of instances.

Contrastive learning approaches have made great strides with two ideas: invariant mapping \cite{hadsell2006dimensionality} and instance discrimination \cite{wu2018unsupervised}. That is, the learned representation should be {\bf 1)} stable for certain transformed versions of an instance, and {\bf 2)} distinctive for different instances. Both aspects can be formulated without labels, and the feature learned appears to automatically capture semantic similarity, as benchmarked by downstream classification on standard datasets such as CIFAR100 and ImageNet \cite{chen2020simple}. However, these datasets are curated with distinctive and class-balanced instances, whereas natural data could be highly correlated within the class  (e.g., repeats) and long-tail distributed across classes.

\figteaser{!t}

Natural between-instance similarity demands instance grouping not instance discrimination, where {\it all the instances are presumed different}.  Consequently, feature learning by instance discrimination is unstable and under-performing without instance grouping, whereas instance grouping based on the feature learned without instance discrimination is easily trapped into degeneracy.   Ad-hoc tricks \cite{caron2018deep,caron2019unsupervised} and mutual information maximization with a uniform class distribution prior \cite{ji19invariant}
have been used to prevent feature degeneracy.

We propose to discover and integrate between-instance similarity into contrastive learning, not directly by instance grouping, 
e.g., by imposing group-level discrimination as DeepCluster \cite{caron2018deep,caron2019unsupervised}
or by regulating instance-level discrimination based on the grouping outcome as Local Aggregation (LA) \cite{zhuang2019local},
but by imposing cross-level discrimination (CLD) between instances and local instance groups.

Contrastive learning is built upon dual forces of attraction and repulsion \cite{hadsell2006dimensionality}.  Existing methods generally assume repulsion between different instances and attraction within {\it known groupings} of instances, e.g., between augmented views of the same data instances \cite{wu2018unsupervised,zhuang2019local,he2020momentum}, or between data captured from different times, views, or modalities of the same physical instances \cite{oord2018representation, bachman2019amdim,tschannen2019mutual,tian2019contrastive}.  

Feature learning with between-instance similarity calls for attraction within {\it unknown groupings}, not the universal between-instance repulsion   (\fig{teaser}a).  An chicken-and-egg challenge is to discover such groupings for feature learning while the feature for the groupings is still to be developed.

Our key insight is that grouping could result from not just attraction, but also common repulsion.  While invariant mapping is achieved by within-instance similarity from attraction across augmented views, between-instance similarity can emerge from repulsion against common instance groups, the centroids of which are more stable in the developing feature space.
That is, to discover the most discriminative feature that also respects natural instance grouping, we desire each instance to attract the closest group related by augmentation and repel groups of other instances that are far from it.  

In our approach (\fig{teaser}b), 
between-instance similarity, unknown {\it a priori}, is not captured directly as attraction between instances, but by more likely common attraction and repulsion between each instance and instance group centroids.  By pulling an instance towards and pushing it against more stable instance groups, {\it similar} instances get mapped closer in the feature space.  To effect both grouping and discrimination objectives on feature learning, we also impose them on features separately derived from a shared representation.

Such an interplay between attraction and repulsion has been utilized to model perceptual popout \cite{yu2001understanding,bernardis2010finding}, as well as simultaneous image segmentation and depth segregation \cite{yu2001segmentation,maire2011object}.   However, those works are prior to deep learning and aim at grouping pixels based on certain fixed pixel-level feature such as edges, whereas our work aims at learning the image-level feature discriminatively.

We add CLD to popular state-of-the-art (SOTA) unsupervised feature learning approaches  (\fig{teaser}c), e.g.,
NPID \cite{wu2018unsupervised}, 
MoCo \cite{he2020momentum},
InfoMin \cite{tian2020makes} (all three based on instance discrimination),
and BYOL \cite{grill2020bootstrap} (focusing only on invariant mapping without instance discrimination).   
CLD delivers a significant performance boost not only on highly correlated, long-tail, and balanced datasets, but also on all the self-supervision, semi-supervision, and transfer learning benchmarks under fair comparison settings \cite{wu2018unsupervised,he2020momentum,openselfsup}.  

Our work makes three major contributions. 
\textbf{1)} We extend unsupervised feature learning to natural data with high correlation and long-tail distributions.
\textbf{2)} We propose cross-level discrimination between instances and local groups, to discover and integrate between-instance similarity into contrastive learning.  We also propose normalized projection heads and unsupervised hyper-parameter tuning.
\textbf{3)} Our experimentation demonstrates that adding CLD to existing methods has an negligible overhead and yet delivers a significant boost.  It  achieves new SOTA on all the benchmarks, and beats MoCo v2 \cite{chen2020improved} and SimCLR \cite{chen2020simple} on every reported performance attained with a much larger compute.

\section{Related Works}

Unsupervised representation learning~\cite{doersch2015unsupervised,pathak2016context,zhang2016colorful,noroozi2016unsupervised,donahue2017adversarial,larsson2017colorization,jenni2018self,gidaris2018unsupervised,zhan2019self} aims to learn features transferable to downstream tasks.  Our work is closely related to contrastive learning and unsupervised feature learning with grouping.

\noindent\textbf{Contrastive learning} maps positive samples closer and negative samples apart in the feature space
\cite{wu2018unsupervised, misra2019self, tian2019contrastive, he2020momentum, chen2020improved, chen2020simple}.
Positive samples come from augmented views of each instance, whereas negative ones come from different instances.   The key distinction among existing methods lies in how these samples are obtained and maintained during learning. 

{\bf Batch methods} ~\cite{chen2020simple} draw samples from the current mini-batch with the same encoder, updated end-to-end with back-propagation.   {\bf Memory-bank methods} ~\cite{wu2018unsupervised, misra2019self} draw samples from a memory bank that stores the prototypes of all the instances computed previously.  {\bf  Hybrid methods} ~\cite{he2020momentum, chen2020improved} encode positive samples by a momentum-updated encoder and maintain negative samples in a queue.

Instance discrimination methods presume distinctive instances.  Their performance drops on natural data that are highly correlated or long-tail distributed,  e.g., consecutive frames in a video,
 or different views of the same instance.  Note that our setting is {\it completely unsupervised} and different from learning representation across views \cite{bachman2019amdim,tschannen2019mutual,tian2019contrastive}:  We have  mixed  data without any object or view labels.

\noindent\textbf{Feature learning with grouping}
exploits natural organization of data
 ~\cite{xie2016unsupervised,yang2016joint,caron2019unsupervised,zhuang2019local}. Unlike self-supervised learning ~\cite{pathak2016context,noroozi2016unsupervised,gidaris2018unsupervised}, it does not require  domain knowledge ~\cite{caron2018deep}. 

Earlier works restrict learning to linear feature transformations.  DisCluster \cite{de2006discriminative,ding2007adaptive} and DisKmeans \cite{ye2008discriminative} iteratively apply K-means to generate cluster labels and then use linear discriminant analysis (LDA) to select the most discriminative subspace.
\cite{yang2010image} applies LDA along with spectral clustering \cite{von2007tutorial}. \cite{nie2011spectral} uses linear regression as a regularization term to handle out-of-sample data in spectral clustering.
 
 Nonlinear feature transformations have also been studied.  \cite{tian2014learning} applies a deep sparse autoencoder to a normalized graph similarity matrix and performs K-means on the latent representation.
\cite{van2009learning} implements t-SNE embedding with a deep neural network.
Deep Embedded Clustering~\cite{xie2016unsupervised} simultaneously learns cluster centroids and feature mapping such that centroid-based soft assignments in the embedding matches a desirable target distribution. 

Recent works 
jointly optimize the feature and the cluster assignment. 
{\bf DeepCluster} ~\cite{caron2018deep, 
caron2019unsupervised} gets pseudo-class labels from global clustering and  applies supervised learning to iteratively fine-tune the model, whereas our CLD incorporates local clustering into contrastive metric learning. 
{\bf Local Aggregation (LA)} ~\cite{zhuang2019local} identifies a local neighbourhood of each instance through clustering, and restricts instance-level discrimination within individual neighbourhoods, whereas CLD looks beyond local neighbourhoods and conducts cross-level instance-group discrimination.
{\bf PCL}  \cite{li2020prototypical} is a con-current work that compares instance features with group centroids which are obtained through global clustering per epoch, whereas our CLD uses local clustering per batch and compares instance-group features within the batch.  Global clusters not only takes more time to compute during training, but conceptually also do not align with classes in downstream tasks.  Empirically, PCL gains much over MoCo but not over MoCo v2 \cite{li2020prototypical}.
{\bf SegSort}~\cite{hwang2019segsort} extends representation learning from classification to segmentation.  It learns a feature per pixel, and assumes that all the pixels in the same region form a cluster in the feature space.
SegSort uses {\it one} common feature and contrasts each {\it pixel} with cluster centroids in the feature from the {\it same}-view, whereas our CLD uses {\it two} separate features and contrasts each {\it image} with cluster centroids in the feature from a {\it different} view.

\noindent{\bf Discussions.} While clustering on a fixed feature is well studied \cite{gan2007data}, clustering with an adapting feature is a tricky model selection problem:
{\bf 1)} Clustering could fall into trivial solutions where most samples are assigned to a single cluster, trapping feature learning into degeneracy ~\cite{caron2018deep}.
{\bf 2)} Without any external supervision, it is unclear how to ensure that the learned feature  captures latent semantics.

Our work combines contrastive learning and grouping in a single framework, by expanding discrimination between instances to that between instances and local groups.  Discrimination prevents feature learning from degeneracy, while grouping improves stability and helps instance-level discrimination see beyond the finest granularity.  With these two aspects integrated, our CLD significantly improves the learned representation for downstream classification.
\def\figcldDiagram#1{
\begin{figure}[#1]
\centering
\imw{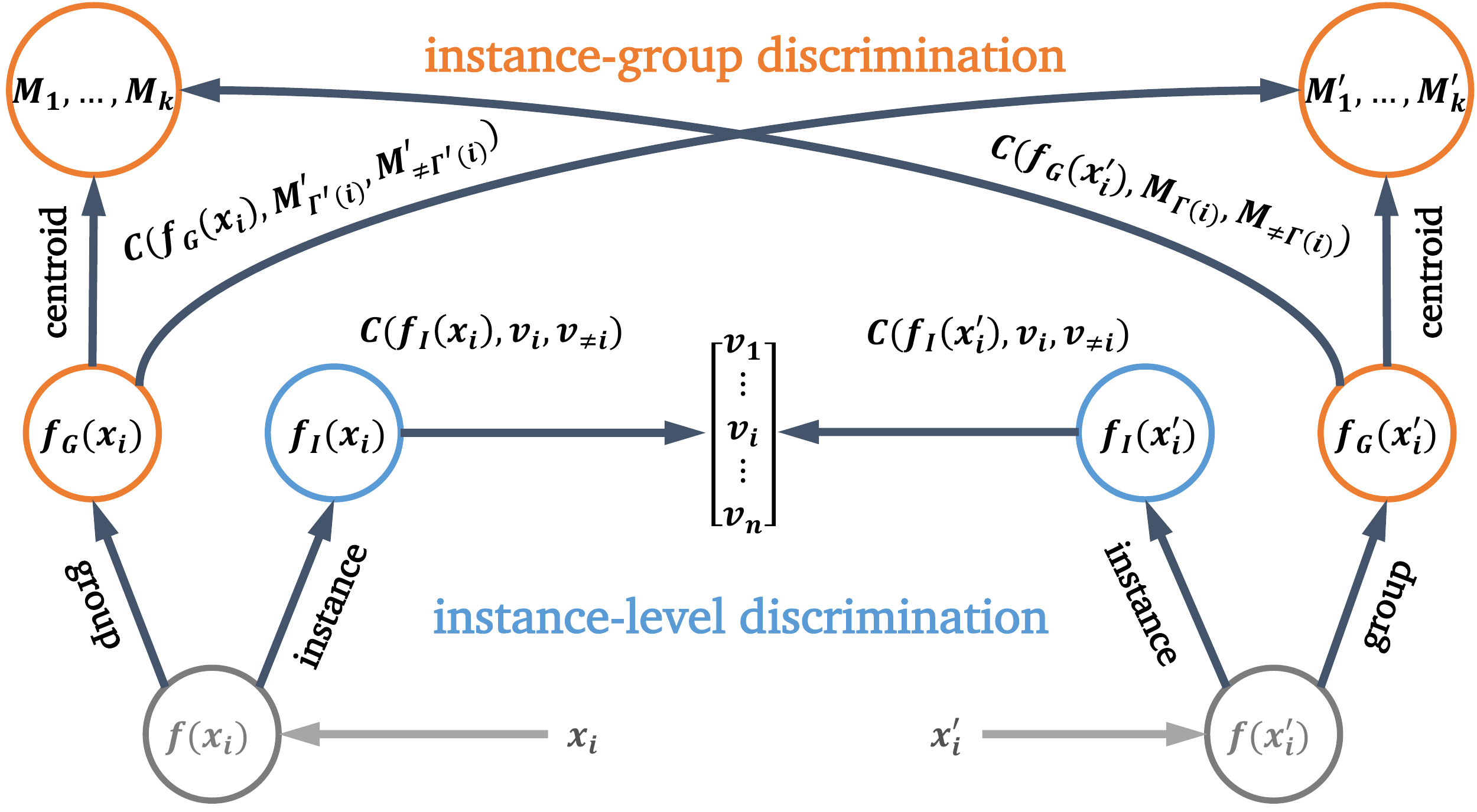}{1.0}
\vspace{-16pt}
\caption{
Method overview.
Our goal is to learn representation $f(x)$ given image $x$ and its alternative view $x'$ from data augmentation.  We fork two branches from $f$: fine-grained \textit{instance branch} $f_I$ and coarse-grained \textit{group branch} $f_G$.  All the computation is mirrored and symmetrical with respect to different views of the same instance.
{\bf 1) Instance Branch}:  We apply contrastive loss (\textit{two bottom $C$'s}) between $f_I(x_i)$ and a global memory bank $\{v_i\}$, which holds the prototype for $x_i$, computed from the average feature of the augmented set of $x_i$.
{\bf 2) Group Branch}:  We perform {\it local} clustering of $f_G(x_i)$ for a batch of instances to find $k$ centroids, $\{M_1,\ldots,M_k\}$, with instance $i$ assigned to centroid $\Gamma(i)$.   Their counterparts in the alternative view are $f_G(x_i')$, $M'$, and $\Gamma'$.
{\bf 3) Cross-Level Discrimination}:
We apply contrastive loss (\textit{two top $C$'s}) between feature $f_G(x_i)$ and centroids $M'$ according to grouping $\Gamma'$, and vice versa for $x_i'$. 
{\bf 4)}
Two similar instances $x_i$ and $x_j$
would be pushed apart by the instance-level contrastive loss but pulled closer by the cross-level contrastive loss, as they repel common negative groups.
Forces from branches $f_I$ and $f_G$ act on their common feature basis $f$, organizing it into one that respects both instance grouping and instance discrimination.
}
\label{fig:cldDiagram}
\end{figure}
}

\section{Learning with Cross-Level Discrimination}


Given $n$ images, we regard instance $x_i$ as a {\it view} obtained by a certain transformation (e.g. cropping) of the $i$-th image.   Let $x_i$ and $x_i'$ denote two different {\it views} of the $i$-th instance.  

\noindent{\bf Contrastive learning}
\cite{hadsell2006dimensionality,wu2018unsupervised,he2020momentum, tian2019contrastive, oord2018representation,chen2020simple} aims to learn a mapping function $f$ such that in the  $f(x)$ feature space, instance $x_i$ is
{\bf 1)} close to positive sample  $x_i'$ {\bf (invariant mapping)}, and
{(\bf 2)} far from negative sample $x_j$ (with $j\!\neq\! i$) of any other instances {\bf (instance discrimination)}.

We model $f$ by a convolutional neural network (CNN) with parameters $\theta$, mapping $x$ onto a $d$-dimensional hypersphere such that $\|f(x)\|\!=\!\!1$.  
Let $f$, $f^+$, $f^-$ denote the feature for an instance and its positive / negative samples respectively.   We optimize $\theta$ by minimizing loss $C$ over all $n$ instances so that $f$ attracts $f^+$ and repels $f^-$.

{\bf instance-centric contrastive loss:}
{\small
\begin{align}
C\left(f_i,f^+_i, f^-_{\neq i}\right) \!=\! 
-\log \frac%
{\exp\frac{<f_i,f_i^+>}{T}}
{\exp\frac{<f_i,f_i^+>}{T}+\sum\limits_{j\neq i}\exp\frac{<f_i,f_j^->}{T}}
\!\!\!
\!\!\!
\end{align}
}%
Temperature $T$ is a hyperparameter regulating what distance is close.   $C$ is the noise contrastive estimation (NCE) \cite{gutmann2012noise} of softmax instance classification loss \cite{wu2018unsupervised}, and it can be viewed as maximizing a lower bound of mutual information (MI) between samples of the same instances \cite{paninski2003estimation,hadsell2006dimensionality, oord2018representation}.

\figcldDiagram{!t}

\noindent{\bf Implementation of $(f_i,f^+_i, f^-_{\neq i})$ during training.}
For sample $x_i$, the self feature is $f_i\!=\!f(x_i)$, whereas positive feature $f_i^+$ and negative feature $f_{\neq i}^-$ come from a memory bank $v$ that
holds the representative feature for $\{x_i\}_{i=1}^n$.  It is computed as the average feature of all the augmented versions of $x_i$ seen so far \cite{wu2018unsupervised, chen2020simple}.  It could also be encoded by a parametric model as in MoCo \cite{he2020momentum}.   Existing methods apply $C$ at the instance level, between instance feature $f_I$ and its average $v$:  $C\left(f_I(x_i),v_i,v_{\neq i}\right)$ (\fig{cldDiagram} instance branch).

\noindent{\bf Pros and cons of instance-level contrastive learning.}  
Contrastive learning has greatly closed the gap with supervised classification \cite{wu2018unsupervised,oord2018representation, he2020momentum, chen2020simple}. However, there are 4 caveats.
\begin{enumerate}[leftmargin=*,itemsep=-3pt,topsep=2pt]
\item It focuses on within-instance similarity by data augmentation, oblivious of between-instance similarity. 
\item It focuses on discrimination at the finest instance level, oblivious of natural groups which often underlie downstream tasks' discrimination at a coarser semantic level.
\item It presumes distinctive instances, whereas non-curated data could contain repeats, redundant observations of the same instance, and long-tail distributed instances across classes in the downstream task.  For feature $f_i$, its negative features $\{f_i^-\}$ would thus contain highly correlated samples which $f_i$ should ideally attract rather than repel.
\item Each instance has a high   positive/negative imbalance ratio (1 vs. rest);  the more negatives, the larger the signal to noise ratio \cite{poole2019variational}, and the better the performance \cite{hjelm2018learning, tian2019contrastive}.  However, the model also leans towards more instance discrimination than invariant mapping,
reducing robustness.
\end{enumerate}

\noindent{\bf Feature grouping.}  To overcome these caveates, we step beyond individual instances and discover how they might be related.  We acknowledge the natural grouping of instances by finding {\it local} clusters within a batch of samples.  Which specific clustering method to use is not as critical; we apply spherical K-means to the unit-length feature vectors.


Local clustering could be rather noisy, especially at the early stage of learning. Instead of imposing group-level discrimination, we validate local groupings across views and impose consistent discrimination between individual instances and their cross-view local groups.

\noindent{\bf Group branch.} Grouping and discrimination are opposite in nature.  To effect both objectives, we fork two branches (just one FC layer each) from feature $f$: fine-grained instance branch $f_I$ 
and coarse-grained group branch $f_G$ (\fig{cldDiagram}).
We first extract $f_G$ at the instance level in a batch, then compute $k$ local cluster centroids $\{M_1,\ldots,M_k\}$ and assign each instance to its nearest centroid.  Clustering assignment $\Gamma(i)\!=\!j$ means that instance $i$ is assigned to centroid $j$. 

\noindent{\bf Cross-level discrimination.}  Natural groups identified in the group branch allows the expansion of positive samples from augmented versions of an individual instance to like-kind {\it other} instances.  We also expand negative samples from other instances to groups of their like-kind instances.  We apply {\it local} (i.e., batch-wise) contrastive loss across views between instance feature $f_G(x_i')$ and group centroids $M$, i.e., ${C\left(f_G(x_i'), M_{\Gamma(i)}, M_{\neq\Gamma(i)}\right)}$ and vice versa for $f_G(x_i)$ (\fig{cldDiagram}).  Intuitively, if local clustering $\Gamma$ separates $\{x_i\}$ well, when $x_i$ is replaced by its alternative view $x_i'$, it should still be close to $x_i$'s centroid $M_{\Gamma(i)}$ and far from other centroids $M_{\neq\Gamma(i)}$.
That is, instances and their local clusters should retain their grouping relationships across views.

\noindent
Comparisons across levels, instances, views are beneficial:
\begin{enumerate}[leftmargin=*,itemsep=-3pt,topsep=3pt]
\item For instances clustered in the same group, instance feature $f_G(x_i)$ and $f_G(x_j)$ would be attracted to the same group centroid $M$ or $M'$ and are thus drawn closer.
\item For similar instances $x_i$ and $x_j$ not in the same cluster, they likely repel common group centroids, thereby pulling instance features $f_G(x_i)$ and $f_G(x_j)$ closer.
\item CLD discriminates at 
instance {\it and} group levels, more in line with coarser discrimination at downstream tasks.
\item Comparisons between $f_G$ and $M$ not only avoid direct repulsion between similar instances, but also greatly improves the positive/negative ratio for invariant mapping. For example, the ratio on ImageNet is $\frac{1}{4096}$ for
NPID \cite{wu2018unsupervised}'s set-wise NCE vs. $\frac{1}{255}$ for CLD's batch-wise NCE.
\item Cross-view comparisons between $x_i$ and $x_i'$ focus the model more on invariant mapping.
\end{enumerate}

\noindent{\bf Probabilistic interpretation of CLD.}  Our CLD objective can be understood as minimizing the cross entropy between hard clustering assignment $p_{ij}$  (as {\it ground-truth})  based on $f_G(x_i)$  and soft assignment  $q_{ij}$ predicted from $f_G(x_i')$ in a different view.  Since $p_{ij}\!=\!1$ only when $j\!=\!\Gamma(i)$, we have a loss that validates local groupings across different views:
\begin{align}
-E_p[\log q]
&=\sum_{i} C(f_G(x_i'), M_{\Gamma(i)},  M_{\neq\Gamma(i)}; T_G).
\end{align}
\noindent{\bf Total contrastive learning loss.}
We add CLD to instance discrimination (with temperatures $T_I$, $T_G$, weight $\lambda$) in symmetrical terms over views $x_i$ and  $x_i'$:
{\scriptsize{
\begin{align*}
&L(f;T_I,T_G,\lambda) \!=\!\sum_{i} \!
  \underbrace{C(f_I(x_i),v_i,v_{\neq i};T_I) \!+\!
  C(f_I(x_i'),v_i,v_{\neq i};T_I)}_{\text{instance-level discrimination}} \\
&\!+\!\lambda\sum_{i} \!
  \underbrace{C(f_G(x_i'), M_{\Gamma(i)}, M_{\neq\Gamma(i)};T_G) \!+\!
  C(f_G(x_i), M'_{\Gamma'(i)}, M'_{\Gamma'(i)};T_G)}_{\text{cross-level discrimination}}
\end{align*}}}
We analyze why two feature branches are better than one branch, where $f_I\!=\!f_G$ and $M$ is simply the group centroids of $f_I(x_i)$ or $v$.  In that case, while the instance discriminiation term would repel $x_i$ against any other instances $\{x_j\}$,
the CLD term would make $x_i$ attract {\it some other} instances $\{x_j\}$ in the same group of $x_i$ through their group centroid.  Minimizing the two terms would lead to opposite effects no matter what the local clustering is.  Basing instance feature $f_I$ and group feature $f_G$ as separate branches off feature $f$ would force $f$ to be discriminative enough for the instance branch yet loosely similar enough for the group branch.


\noindent{\bf Normalized projection head.} 
Existing methods derive instance feature $f_I(x)$ by mapping the latent feature $f(x)$ onto a unit hypersphere with  first a projection head and then normalization. 
NPID \cite{wu2018unsupervised} and MoCo \cite{he2020momentum} use one FC layer as a linear projection head. 
MoCo v2 \cite{chen2020improved}, SimCLR \cite{chen2020simple}, and BYOL \cite{grill2020bootstrap} use a multi-layer perceptron (MLP) head; it is better for large datasets and worse for small datasets.

We propose to normalize both the FC layer weights  $W$ and the shared feature vector $f$ so that projecting $f$ onto $W$ simply calculates their cosine similarity.
The $t$-th component of normalized feature $N(x_i)$ (where $N\!=\!f_I$ or $N\!=\!f_G$) is:
\begin{align}
N_t(x_i)
= <\frac{W_t}{\|W_t\|},\frac{f(x_i)}{\| f(x_i)\|}>.
\end{align}
Normalized linear (NormLinear) or MLP (NormMLP) projection heads bring additional gains to CLD.  Empirically, they help reduce feature variance from data augmentation.


\def\figHighCorA#1{
\begin{figure}[#1]\centering
\tb{@{}lr@{}}{1}{
\tb{c}{0}{
\includegraphics[width=0.26\textwidth]{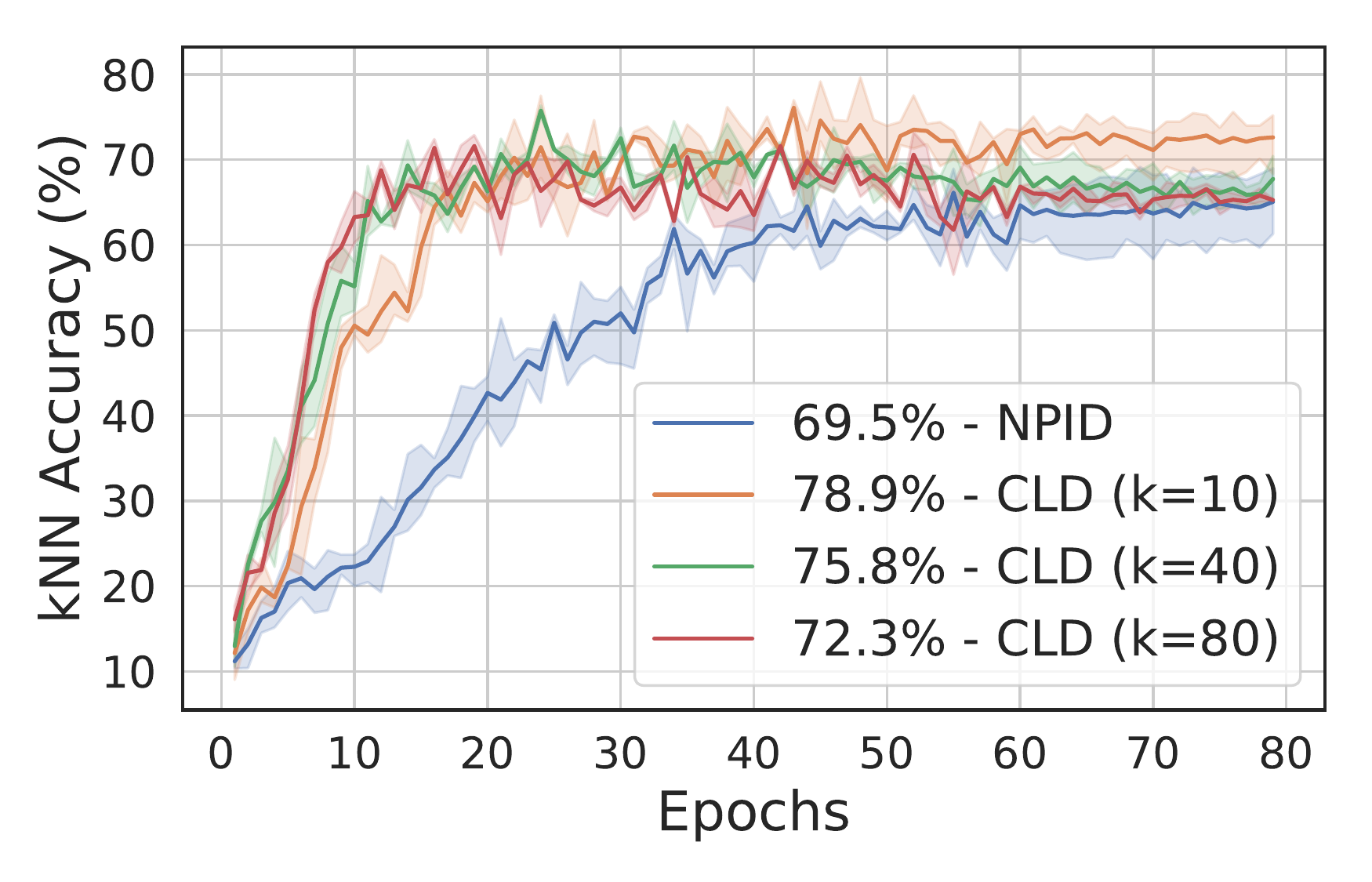}\\[-5pt]
}\hspace{-3pt} & 
\tablestyle{2.0pt}{0.98}
\tb{@{}l|l@{}}{1}{
\shline
Kitchen- & kNN \\
HC & Accuracy \\
\shline
NPID &  69.5 \\
NPID + CLD & 78.9 {\color{green(pigment)} \scriptsize \bf (+9.4)} \\
\hline
MoCo &  76.1 \\
MoCo + CLD & {\bf{81.6}} {\color{green(pigment)} \scriptsize \bf (+5.5)} \\
\shline
}
}\vspace{-10pt}
\caption{{\bf Left}: CLD is more accurate and fast converging than NPID on Kitchen-HC, esp. when the number of groups is closer to the number of classes 11. The average top-1 kNN accuracy of 5 runs is reported.
{\bf Right}: CLD outperforms NPID or MoCo on \textbf{high correlation dataset} Kitchen-HC. }
\label{fig:HighCorA}
\end{figure}
}

\def\figHighCorB#1{
\begin{figure}[#1]
\centering
\includegraphics[width=0.44\textwidth]{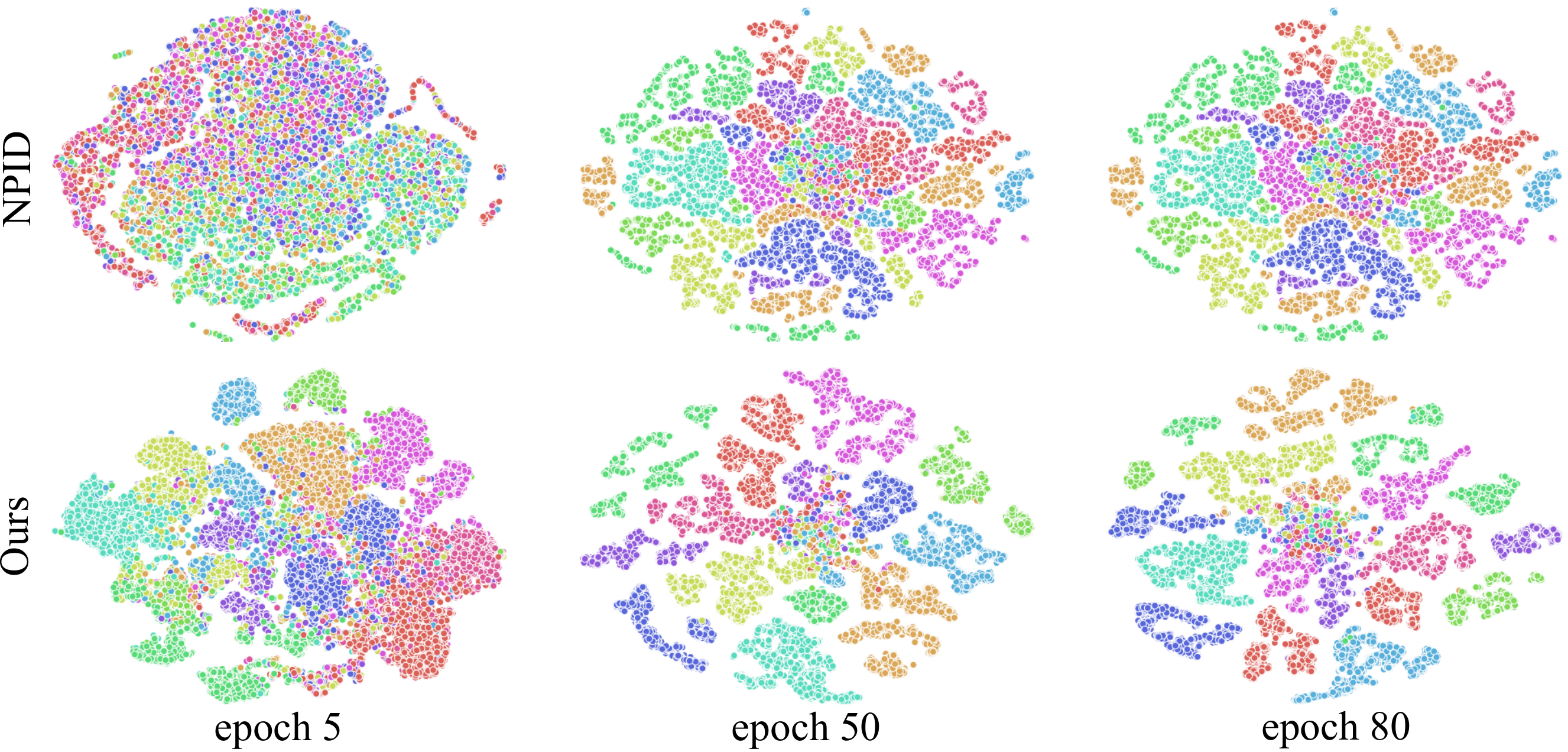}\vspace{-10pt}
\caption{CLD has earlier and better separation between classes (indicated by the dot color) than NPID in the \textbf{t-SNE visualization} of instance feature $f_I(x_i)$ on Kitchen-HC.}
\label{fig:HighCorB}
\end{figure}
}

\def\tabLongtail#1{
\begin{table}[#1]
\tablestyle{2.0pt}{0.98}
\begin{tabular}{l|cc|cc|lllcc}
\shline
\multirow{2}{*}{} & \multicolumn{2}{c|}{CIFAR10-LT} & \multicolumn{2}{c|}{CIFAR100-LT} & \multicolumn{3}{c}{ImageNet-LT}\\ 
\cline{2-8}
 & top1 & top5 & top1 & top5 & many/med/few & top1 & top5 \\ [.1em]
\shline
\multicolumn{8}{l}{\textit{Unsupervised}} \\
\hline
NPID \cite{wu2018unsupervised} & 32.3 & 74.8 & 10.2 & 29.8 & 47.5/21.3/6.6 & 29.5 & 51.1 \\
NPID + CLD & 41.1 & 78.9 & 21.7 & 44.3 & 52.4/25.0/8.3 & 32.7 & 55.6 \\
\it vs. baseline & \color{green(pigment)} \footnotesize \bf +8.8 &\color{green(pigment)} \footnotesize \bf +4.1 &\color{green(pigment)} \footnotesize \bf +11.5 &\color{green(pigment)} \footnotesize \bf +14.5 &\color{green(pigment)} \footnotesize \bf {+4.9}/{+3.7}/{+1.7} &\color{green(pigment)} \footnotesize \bf {+3.2} &\color{green(pigment)} \footnotesize \bf {+4.5}\\
\hline
MoCo \cite{he2020momentum} & 34.2 & 76.7 & 19.7 & 42.6 & 48.1/21.3/6.9 & 29.9 & 51.8 \\
MoCo + CLD & \bf{43.1} & \bf{80.4} & \bf{25.4} & \bf{50.0} & \bf{53.1}/\bf{24.9}/\bf{9.4} & \bf{33.3} & \bf{57.3} \\
\it vs. baseline & \color{green(pigment)} \footnotesize \bf +8.9 &\color{green(pigment)} \footnotesize \bf +3.7 &\color{green(pigment)} \footnotesize \bf +5.7 &\color{green(pigment)} \footnotesize \bf +7.4 &\color{green(pigment)} \footnotesize \bf {+5.0}/{+3.6}/{+2.5} &\color{green(pigment)} \footnotesize \bf +3.4 &\color{green(pigment)} \footnotesize \bf +5.5 \\[.1em]
\hline
\multicolumn{8}{l}{\textit{Supervised}} \\
\hline
CE & - & - & - & - & 40.9/10.7/0.4 & 20.9 & - \\
OLTR \cite{liu2019large} & - & - & - & - & 43.2/35.1/18.5 & 35.6 & - \\ [.1em]
\shline
\end{tabular}\vspace{-8pt}
\caption{CLD outperforms unsupervised baselines on \textbf{long-tailed datasets}, approaching supervised cross-entropy (CE) and OLTR \cite{liu2019large}. The kNN (linear) classifiers are used for CIFAR (ImageNet-LT). CLD is significantly better than supervised CE on many-shot ($100+$), medium-shot ($[20,100)$), few-shot ($20-$), and gets close to OLTR.  }
\label{table:long_tail_cifar}
\end{table}
}

\def\tabSmallOnly#1{
\begin{table}[#1]
\tablestyle{2.0pt}{0.98}
\setlength{\tabcolsep}{5pt}
\begin{tabular}{l|c|c|c|c}
\shline
kNN accuracies & STL10 & CIFAR10 & CIFAR100 & ImageNet100 \\ [.1em]
\shline
DeepCluster & - & \ 67.6 & - & - \\
Exemplar \cite{dosovitskiy2015discriminative}  & 79.3 & 76.5 & - & - \\
Inv. Spread \cite{ye2019unsupervised} & 81.6 & 83.6 & - & - \\
CMC \cite{tian2019contrastive} & - & - & - & 79.2 \\
\hline
NPID \cite{wu2018unsupervised} & 79.1 & 80.8 & 51.6 & 75.3 \\
NPID + CLD & 83.6 & 86.7 & 57.5 & 79.7 \\
\it vs. baseline & \color{green(pigment)} \footnotesize \bf +4.5 &\color{green(pigment)} \footnotesize \bf +5.9 &\color{green(pigment)} \footnotesize \bf +5.9 &\color{green(pigment)} \footnotesize \bf +3.6 \\
\hline
MoCo \cite{he2020momentum} & 80.8 & 82.1 & 53.1 & 76.6 \\
MoCo + CLD & \bf{84.3} & \bf{87.5} & \bf{58.1} & \bf{81.5} \\
\it vs. baseline & \color{green(pigment)} \footnotesize \bf +3.5 &\color{green(pigment)} \footnotesize \bf +5.4 &\color{green(pigment)} \footnotesize \bf +5.0 &\color{green(pigment)} \footnotesize \bf +4.9 \\
\hline
BYOL \cite{grill2020bootstrap} &-&-&-& 75.8 \\
BYOL + CLD &-&-&-& 81.1 \\
\it vs. baseline &-&-&-& \color{green(pigment)} \footnotesize \bf +4.7 \\
\shline
\end{tabular}\vspace{-8pt}
\caption{\textbf{On self-supervised learning on small/medium-sized benchmarks}: STL10, CIFAR10, CIFAR100 and ImageNet-100, CLD delivers consistent gains as an add-on to various methods which use either standard contrastive loss (e.g. MoCo \cite{he2020momentum}) or without negative pairs (e.g. BYOL \cite{grill2020bootstrap}).  On ImageNet-100, we use our re-implemented code for baselines as they are better than those in CMC \cite{tian2019contrastive}. All baselines and their CLD add-on's are optimized with the same training recipe for fair comparisons.  For small- and medium-sized datasets, the nonlinear multi-layer perceptron (MLP) head performs worse than a linear projection head.}
\label{table:small_datasets}
\end{table}
}

\def\tabSplitsGroups#1{
\begin{table}[#1]
\begin{minipage}[t][][b]{.30\textwidth}
    \tablestyle{4.0pt}{1.05}
    \setlength{\tabcolsep}{5pt}
    \begin{tabular}{ll|c|c|c}
    \shline
    \multicolumn{2}{l|}{CIFAR10} & retrieval & $\text{NMI}$ & kNN \\ [.1em]
    \shline
    \multirow{1}{*}{NPID}  & $f_I$ & 75.1 & 57.7 & 80.8 \\
    \hline
    \multirow{2}{*}{CLD} & $f_I$ & \bf 78.6 & 63.5 &\bf 86.7 \\
     & $f_G$ & 75.6 &\bf 69.0 & 81.4 \\
    \hline
    \multicolumn{2}{l}{CIFAR100} \\ [.1em]
    \hline
    \multirow{1}{*}{NPID}  & $f_I$ & 48.7 & 36.1 & 51.6 \\
    \hline
    \multirow{2}{*}{CLD} & $f_I$ &\bf 50.2 & 43.8 &\bf 57.5 \\
     & $f_G$ & 48.8 &\bf 49.4 & 51.8 \\
    \shline
    \end{tabular}\vspace{-8pt}
    \caption{The \textbf{feature quality} of $f_I$ and $f_G$ evaluated by retrieval, normalized mutual information and kNN.}
    \label{table:branch}
\end{minipage}\hspace{3pt}
\begin{minipage}[t][][b]{.16\textwidth}
    \tablestyle{4.0pt}{1.09}
    \setlength{\tabcolsep}{5pt}
    \begin{tabular}{l|c}
        \shline
        \# groups & top-1 \\
        \shline
        baseline & 53.1 \\
        \hline
        10 & 55.2 \\
        20 & 55.4 \\
        60 & 56.7 \\
        80 & 57.4 \\
        100 & 57.7 \\
        128 & 58.1 \\
        \shline
        \end{tabular}\vspace{-8pt}
        \caption{\textbf{\#groups vs. Accuracy} on CIFAR100 for CLD.}
    \label{table:groups}
\end{minipage}
\end{table}
}

\def\tabImageNet#1{
\begin{table}[#1]
\tablestyle{2.0pt}{0.98}
\setlength{\tabcolsep}{5pt}
\begin{tabular}{l|r|r|r|c}
\shline
Methods & Architecture & 
\#epoch & 
\#GPU &
top-1 \\
\shline
NPID \cite{wu2018unsupervised} & R50-Linear (24M) & 200 & 8 & 56.5 \\
w/ CLD & R50-Linear (24M) & 200 & 8 & 60.6 \\
\hline
MoCo \cite{he2020momentum} & R50-Linear (24M) & 200 & 8 & 60.6 \\
w/ CLD & R50-Linear (24M) & 200 & 8 & 63.4 \\
w/ CLD & R50-NormLinear (24M) & 200 & 8 & 63.8 \\
\hline
MoCo v2 \cite{chen2020improved} & R50-MLP (28M) & 200 & 8 & 67.5 \\
w/ CLD & R50-MLP (28M) & 200 & 8 & 69.2 \\
w/ CLD & R50-NormMLP (28M) & 200 & 8 & 70.0 \\
\hline
BYOL$^\dagger$ \cite{grill2020bootstrap} & R50-MLP (28M) & 100 & \cellcolor{gray!20}128 & 66.5 \\
w/ CLD$^\ddagger$ & R50-NormMLP (28M) & 100 & 8 & 69.1 \\
\hline
InfoMin \cite{tian2020makes} & R50-MLP (28M) & 100 & 8 & 67.4 \\
w/ CLD & R50-MLP (28M) & 100 & 8 & 69.5 \\
w/ CLD & R50-NormMLP (28M) & 100 & 8 & 70.1 \\
\hline
InfoMin \cite{tian2020makes} & R50-MLP (28M) & 200 & 8 & 70.1 \\
w/ CLD & R50-MLP (28M) & 200 & 8 & 70.6 \\
w/ CLD & R50-NormMLP (28M) & 200 & 8 & \bf{71.5} \\
\hline\hline
SimCLR$^\dagger$ \cite{chen2020simple} & R50-MLP (28M) & 100 & \cellcolor{gray!20}128 & 66.5\\
SwAV$^\dagger$ \cite{caron2020unsupervised} & R50-MLP (28M) & 100 & \cellcolor{gray!20}128 & 66.5 \\
BYOL$^\dagger$ \cite{grill2020bootstrap} & R50-MLP (28M) & 100 & \cellcolor{gray!20}128 & 66.5 \\
SimSiam$^\dagger$ \cite{chen2020exploring} & R50-MLP (28M) & 100 & 8 & 68.1 \\
\hline
SimCLR \cite{chen2020simple} & R50-MLP (28M) & 200 & 8 & 61.9\\
SimCLR$^\dagger$ \cite{chen2020simple} & R50-MLP (28M) & 200 & \cellcolor{gray!20}128 & 68.3\\
SwAV$^\dagger$ \cite{caron2020unsupervised} & R50-MLP (28M) & 200 & \cellcolor{gray!20}128 & 69.1 \\
BYOL$^\dagger$ \cite{grill2020bootstrap} & R50-MLP (28M) & 200 & \cellcolor{gray!20}128 & 70.6 \\
MoCo v2 \cite{chen2020improved} & R50-MLP (28M) & 200 & 8 & 67.5 \\
SimSiam$^\dagger$ \cite{chen2020exploring} & R50-MLP (28M) & 200 & 8 & 70.0 \\
\hline
PIRL \cite{misra2019self} & R50-Linear (24M) & \cellcolor{gray!20}800 & \cellcolor{gray!20}32 & 63.6\\
CMC \cite{tian2019contrastive} & \cellcolor{gray!20}$\text{R50}_\text{{L+ab}}$-Linear (47M) & \cellcolor{gray!20}280 & 8 & 64.1 \\
CPC v2 \cite{henaff2019data} & \cellcolor{gray!20}R170$^{\*}$-Linear (303M) & 200 & \cellcolor{gray!20}32 & 65.9 \\
SimCLR \cite{chen2020simple} & R50-MLP (28M) & \cellcolor{gray!20}800 & \cellcolor{gray!20}128 & 69.3 \\
MoCo v2 \cite{chen2020improved} & R50-MLP (28M) & \cellcolor{gray!20}800 & 8 & 71.1 \\
SwAV \cite{caron2020unsupervised} & R50-MLP (28M) & \cellcolor{gray!20}400 & \cellcolor{gray!20}128 & 70.1 \\
SimSiam$^\dagger$ \cite{chen2020exploring} & R50-MLP (28M) & \cellcolor{gray!20}800 & 8 & 71.3 \\
\shline
\end{tabular}\vspace{-8pt}
\caption{
{\bf On self-supervised learning on ImageNet}, our CLD and NormMLP can be added to improve existing methods and achieve SOTA under 100-/200-epoch pre-training settings.
Note that our experiments with CLD are conducted with 8 RTX 2080Ti GPUs, whereas PIRL, SimCLR, BYOL and SwAV require batch size 4,096 and 128/512 GPUs/TPUs for their original reported performance.
All the results follow the standard linear evaluation protocol as used in \cite{wu2018unsupervised,he2020momentum,chen2020improved,tian2020makes}, except those marked by $\dagger$ (all copied from \cite{chen2020exploring}): The linear classifier training of SwAV \cite{caron2020unsupervised}, BYOL \cite{grill2020bootstrap} and SimSiam \cite{chen2020exploring} uses base $lr\!=\! 0.02$ with a cosine decay scheduler, batch size 4096 with a LARS optimizer, giving  these methods about 1\% additional gain \cite{chen2020exploring}.
All the baseline results are from either their original papers or \cite{chen2020exploring}.
For BYOL$+$CLD results marked by $\ddagger$, the target network is updated once every 16 steps and uses batch size 256.
}
\label{table:imagenet}
\end{table}\vspace{0mm}
}


\def\tabSemi#1{
\begin{table}[#1]
    \tablestyle{2.0pt}{0.98}
    \centering
    \setlength{\tabcolsep}{13pt}
    \begin{tabular}{l|c|cc}
        \shline
       \multirow{2}{*}{Methods} & \multirow{2}{*}{Model} & \multicolumn{2}{c}{Label fraction} \\
         & & 1\% & 10\% \\
        \shline
        random initialization & ResNet50 & 1.6 & 21.8 \\
        \hline
        rotation \cite{gidaris2018unsupervised} & ResNet50 & 19.0 & 53.9 \\
        DeepCluster \cite{caron2018deep} & ResNet50 & 33.4 & 52.9 \\
        NPID \cite{wu2018unsupervised} & ResNet50 & 28.0 & 57.2 \\
        MoCo \cite{he2020momentum} & ResNet50 & 33.2 & 60.1 \\
        SimCLR \cite{chen2020simple} & ResNet50 & 36.3 & 58.5 \\
        MoCo v2 \cite{chen2020improved} & ResNet50 & 38.7 & 61.6 \\
        InfoMin $\dagger$ \cite{tian2020makes} & ResNet50 & 39.7 & 62.3 \\
        \hline
        MoCo v2 + CLD & ResNet50 & 44.4 & 63.6 \\
        InfoMin + CLD & ResNet50 & \bf{45.8} & \bf{64.4} \\
        \it vs. SOTA &ResNet50 & \color{green(pigment)} \footnotesize \bf +6.1 & \color{green(pigment)} \footnotesize \bf +2.1 \\
        \shline
    \end{tabular}\vspace{-8pt}
    \caption{Top-1 accuracy of {\bf{semi-supervised learning}} (1\% and 10\% label fractions) on ImageNet. CLD greatly improves SOTA. Baselines and CLD follow training recipes of OpenSelfSup benchmark \cite{openselfsup} for fair comparisons, and apply the best performing hyper-parameter setting for each method. $\dagger$ denotes re-implemented results with \cite{openselfsup}.}
    \label{table:semi-super}
  \end{table}
}

\def\tabDetection#1{
\begin{table}[#1]
    \tablestyle{2.0pt}{0.98}
    \centering
    \setlength{\tabcolsep}{11pt}
    \begin{tabular}{l||cc|cc}
    \shline
    \multirow{2}{*}{Methods} & \multicolumn{2}{c|}{VOC07} & \multicolumn{2}{c}{VOC07+12} \\
     & AP${_{50}}$ & AP & AP${_{50}}$ & AP \\
    \shline
    supervised & 74.6&42.4 & 81.3&53.5 \\
    \hline
    JigSaw \cite{goyal2019scaling} & - & - & 82.7 & 53.3 \\
    LocalAgg \cite{zhuang2019local}  & 69.1&- &-&- \\
    MoCo \cite{he2020momentum} & 74.9&46.6 & 81.5&55.9 \\
    MoCo v2 \cite{chen2020improved} & -&-  & 82.0&56.4 \\
    SimCLR \cite{chen2020simple} & 75.2&- & -&- \\
    \hline
    NPID + CLD & 75.7 & 47.2 & 82.0 & 56.4 \\
    MoCo + CLD & 76.8 & 48.3 & 82.4 & 56.7 \\
    MoCo v2 + CLD & 77.6 & 49.3 & 82.7 & 57.0 \\
    InfoMin + CLD & \bf{77.9} & \bf{49.8} & \bf{83.0} & \bf{57.2} \\
    \it vs. SOTA & \color{green(pigment)} \footnotesize \bf +2.7 & \color{green(pigment)} \footnotesize \bf +3.2 & \color{green(pigment)} \footnotesize \bf +1.0 & \color{green(pigment)} \footnotesize \bf +0.8 \\
    \shline
    \end{tabular}\vspace{-8pt}
    \caption{{\bf{Transfer learning}} results on object detection: We fine-tune on Pascal VOC $\textit{trainval07+12}$ or $\textit{trainval07}$, and test on VOC $\textit{test2007}$.  The detector is Faster R-CNN with ResNet50-C4.  MoCo v2 model is pre-trained for 200 epochs.  Note that our model outperforms SOTA methods without using an MLP head. Baseline results are copied from \cite{he2020momentum, chen2020improved}.}
    \label{table:voc_detection} 
\end{table}
}

\def\figRetrievalFull#1{
\begin{figure}[#1]
\centering
\includegraphics[width=\linewidth]{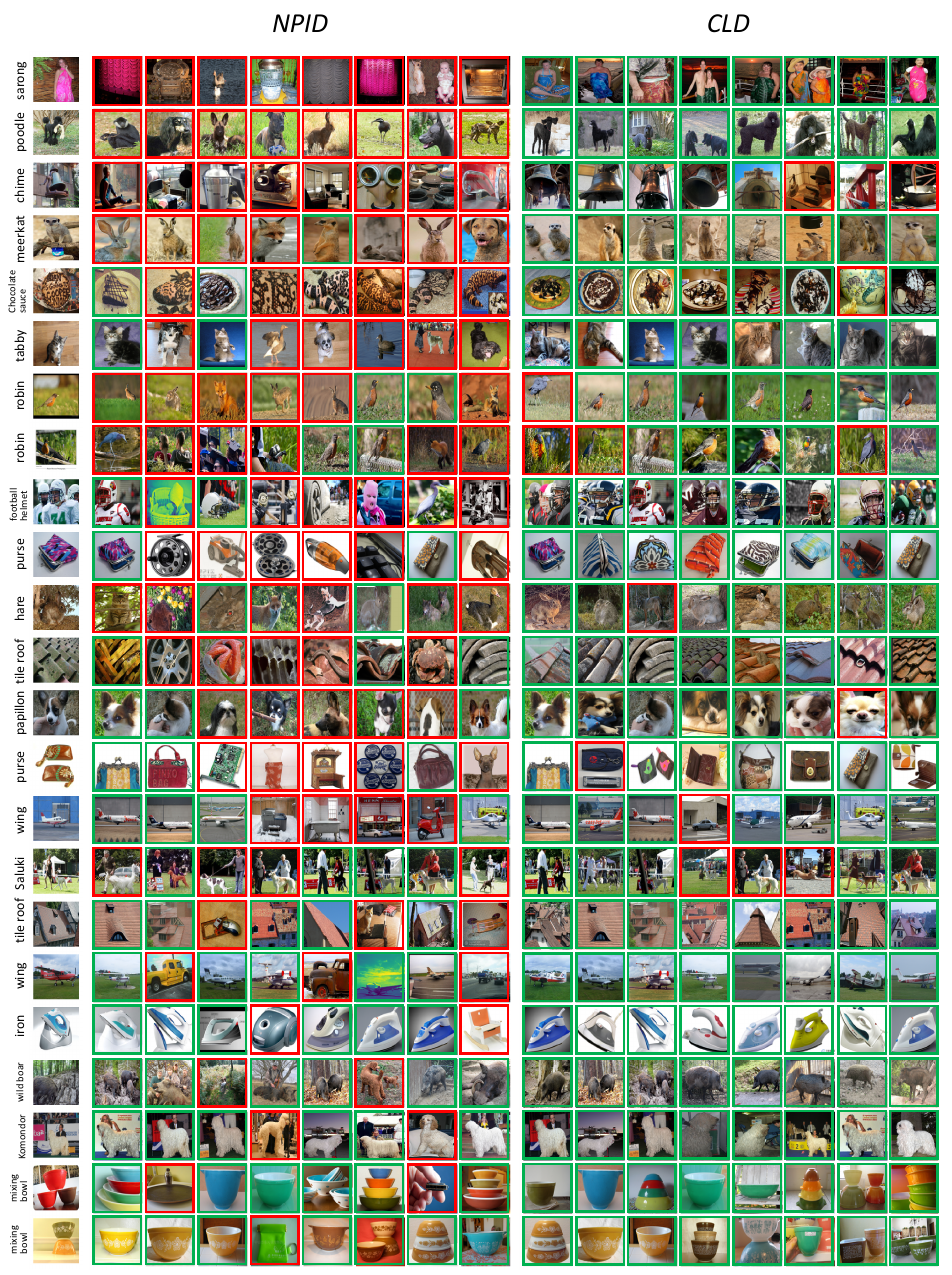}\vspace{-6pt}
\caption{Comparisons of top {{\bf retrieves}} by NPID (Columns 2-9) and CLD (Columns 10-17) according to $f_I$ for the query images (Column 1) from the ImageNet validation set.  The results are sorted by NPID's performance: Retrievals with the same category as the query are outlined in {\color{green} green} and otherwise in {\color{red} red}. NPID seems to be much more sensitive to textural appearance (e.g., Rows 1,4,6,7), first retrieve those with similar textures or colors.  CLD is able to retrieve semantically similar samples.  Our conjecture is that by reducing similar textures into groups, CLD can actually find more informative feature that contrasts between groups. (Zoom in for details)
}
\label{fig:Retrieval}
\end{figure}
}

\def\tabVOC#1{
\begin{table}[#1]
\tablestyle{2.0pt}{0.98}
\begin{tabular}{l||cc|cc}
\shline
\multirow{2}{*}{pre-train methods} & \multicolumn{2}{c|}{VOC07+12} & \multicolumn{2}{c}{VOC07} \\
 & AP$_{50}$ & AP & AP$_{50}$ & AP \\
\hline
random initialization& 60.2 & 33.8  & - & - \\
supervised & 81.3 & 53.5 & 74.6 & 42.4 \\
\hline
Multi-task \cite{doersch2017multi}  & - & -  & 70.5 & - \\
LocalAgg \cite{zhuang2019local}  & - & - & 69.1 & -\\
MoCo \cite{he2020momentum}  & 81.5 & 55.9 & 74.9 & 46.6\\
MoCo v2 \cite{chen2020improved} & 82.0 & 56.4 & - & - \\
SimCLR \cite{chen2020simple} & - & - & 75.2 & - \\
\hline
CLD & 82.0 & 56.4 & 75.7 & 47.2 \\
CLD w/o MLP & 82.4 & 56.7 & 76.8 & 48.3 \\
CLD w MLP & \bf{82.7} & \bf{57.0} & \bf{77.6} & \bf{49.3} \\
\shline
\end{tabular}\vspace{-6pt}
\caption{Transfer learning results on object detection: We fine-tune on Pascal VOC $\textit{trainval07+12}$ or $\textit{trainval07}$, and test on VOC $\textit{test2007}$.  The detector is Faster R-CNN with ResNet50-C4.  MoCo v2 model is trained for 200 epochs with an MLP head.  Note that our model can outperform SOTA methods without using an MLP head. Baseline results are from \cite{he2020momentum, chen2020improved}.}
\label{table:voc_detection}
\end{table}
}

\def\tabLambdaTemp#1{
\begin{table}[#1]\centering
\tb{@{}lr@{}}{2}{
\tablestyle{2.0pt}{0.98}
\tb{l||c|c||c|c}{2}{
\shline
\multirow{3}{*}{Accuracy (\%)} & \multicolumn{2}{c||}{NPID+CLD} & \multicolumn{2}{c}{MoCo+CLD} \\
\cline{2-5}
 & Top-1  & Top-5  & Top-1 & Top-5\\ [.1em]
\hline
$\lambda=0$ (baseline) & 75.3 & 92.4 & 77.6 & 93.8\\ [.1em]
\hline
$\lambda=0.1$ & 78.8 & 94.4 & 80.3 & 95.0\\ [.1em]
$\lambda=0.25$ & \bf{79.7} & \bf{95.1} & \bf{81.7} & \bf{95.7}\\ [.1em]
$\lambda=0.50$ & 78.9 & 94.4 & 80.5 & 95.2\\ [.1em]
$\lambda=1.0$ & 78.8 & 94.5 & 80.1 & 94.8\\ [.1em]
$\lambda=3.0$ & 76.6 & 93.2 & 78.4 & 94.1\\ [.1em]
\shline
}
\tablestyle{2.0pt}{0.98}
\tb{l|c|c|c|c|c|c}{1}{
\shline
$T_I\!=\!T_G$ & 0.07 & 0.1 & 0.2 & 0.3 & 0.4 & 0.5 \\ [.1em]
\hline
Top-1(\%) & 79.3 & 79.6 & 81.7 & 80.7 & 79.4 & 79.0 \\
\shline
}
}\vspace{-6pt}
\caption{
Top-1 and Top-5 linear classification accuracies (\%) on ImageNet-100 with different $\mathbf{\lambda}$'s and temperature $T$'s. $T_I = T_G$ for simplicity.  The backbone network is ResNet-50.
\label{tab:lambdaTemp}
}
\end{table}
}

\def\tabUnsupHyper#1{
\begin{figure}[#1]
    \centering
    \includegraphics[width=1\linewidth]{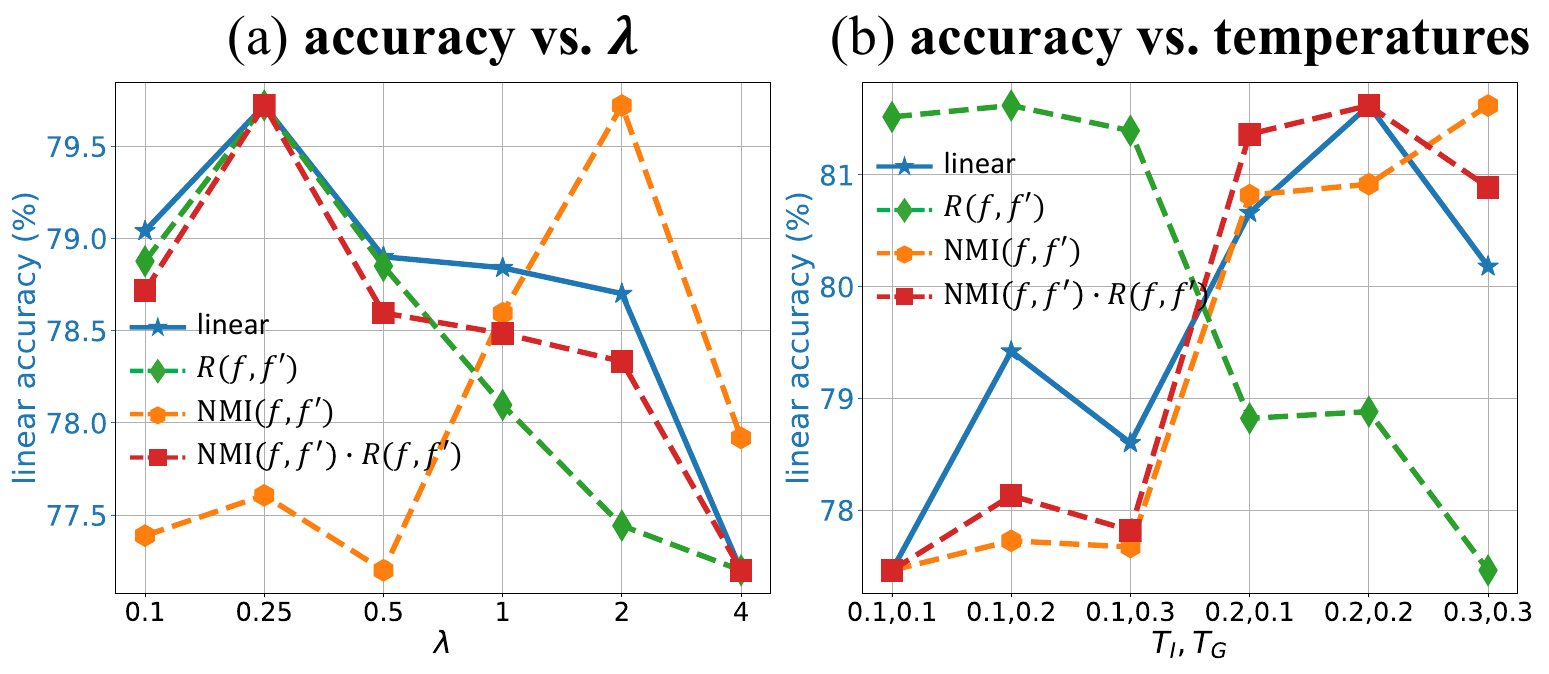}
    \vspace{-24pt}
    \caption{\textbf{Unsupervised hyper-parameter tuning} on ImageNet-100, for weight $\lambda$ (left) and for the temperatures $T_{\text{I}},T_{\text{G}}$ used in CLD (right). Unsupervised evaluation metric $\text{NMI}(f,f')\cdot R(f,f')$ ranks models similarly as supervised linear classification, corroborating our idea that both global mutual information and augmentation-invariant local information are important for downstream performance. Each curve is individually normalized. }    
 \label{fig:unsuper-hyperparam}
\end{figure}
}

\def\figCosSim#1{
\begin{figure}[#1]
\begin{subfigure}{.325\linewidth}
  \centering
  \includegraphics[width=0.99\linewidth]{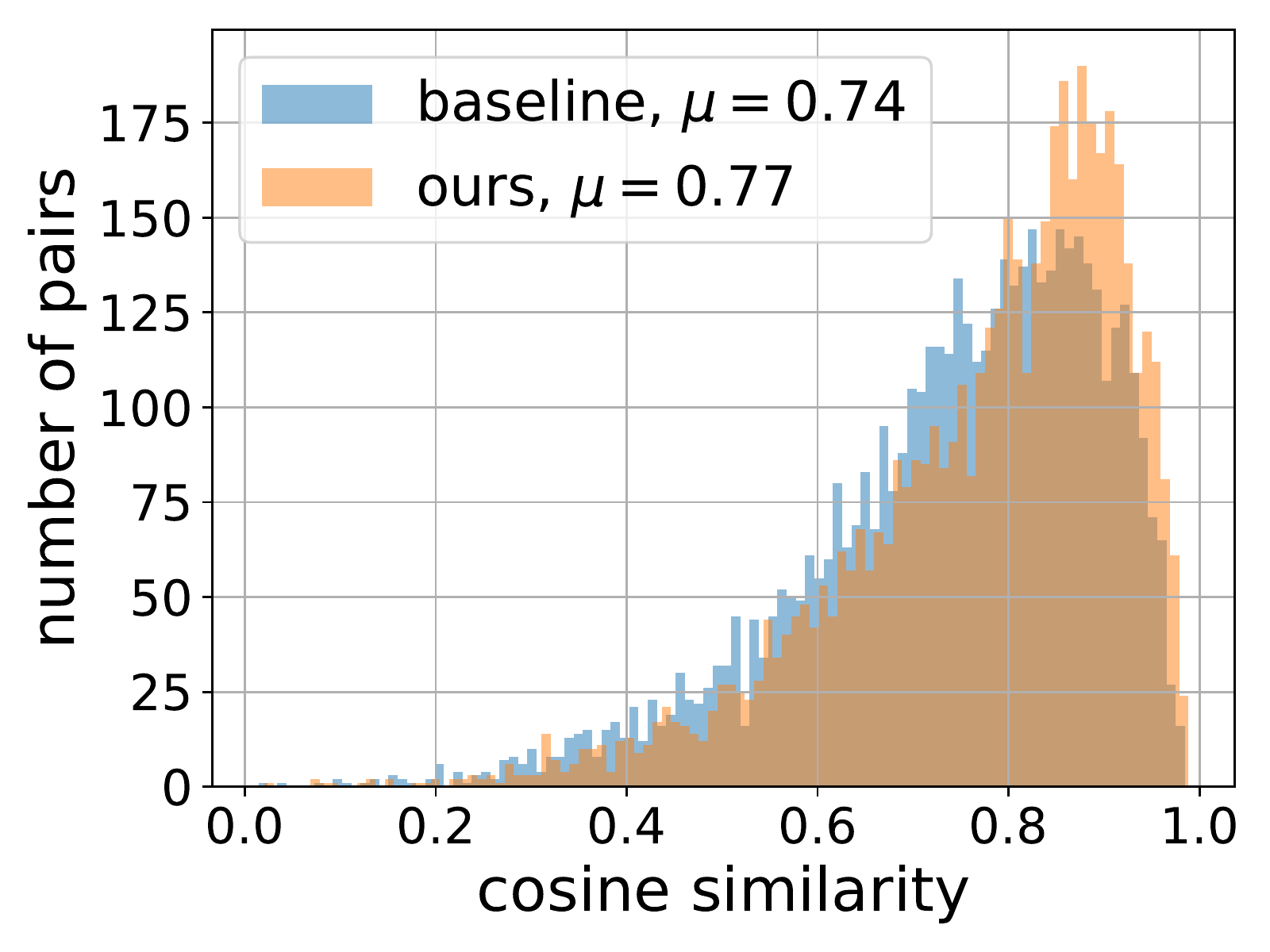}
\end{subfigure}
\begin{subfigure}{.325\linewidth}
  \centering
  \includegraphics[width=0.99\linewidth]{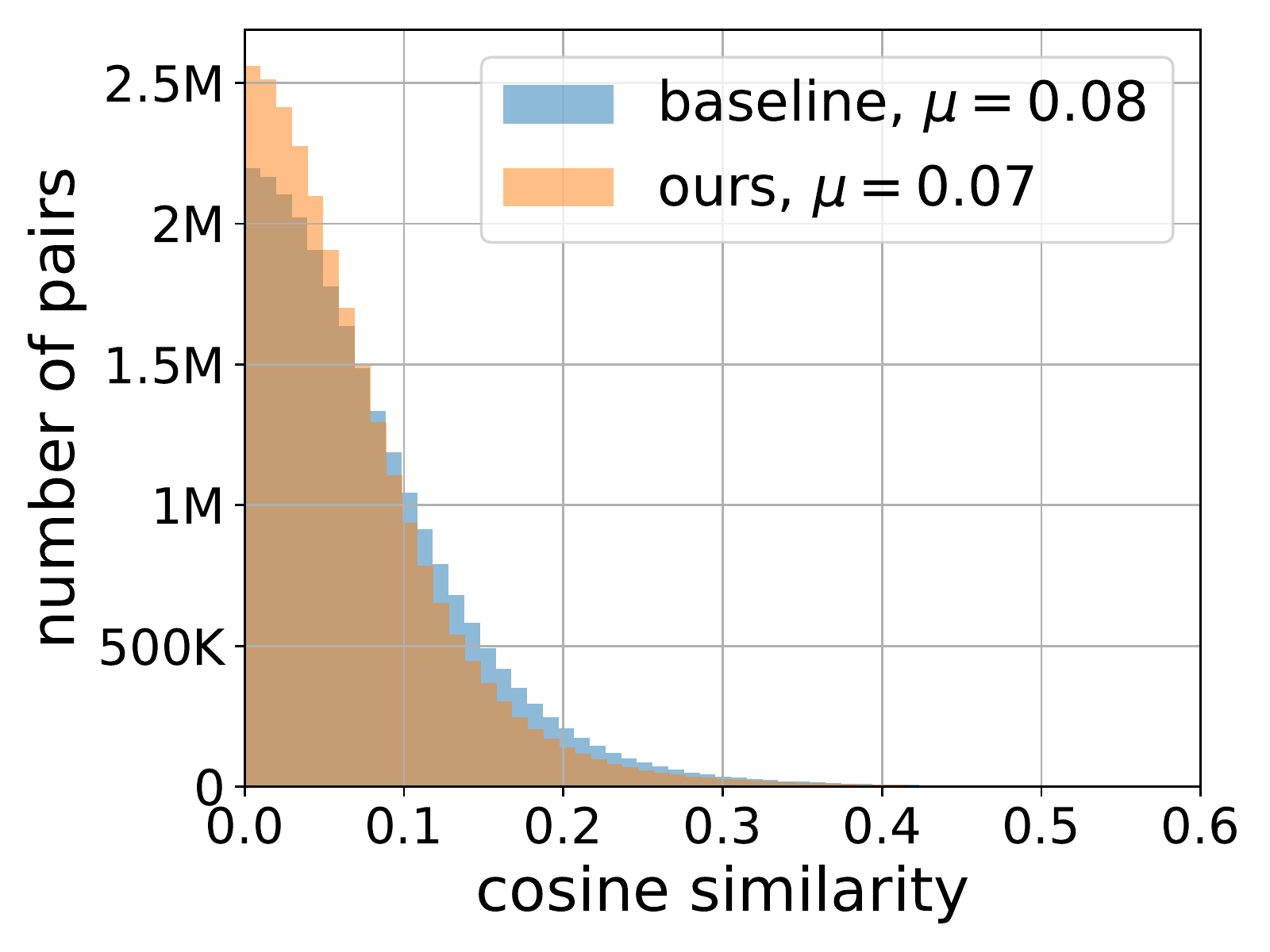}
\end{subfigure}
\begin{subfigure}{.325\linewidth}
  \centering
  \includegraphics[width=0.99\linewidth]{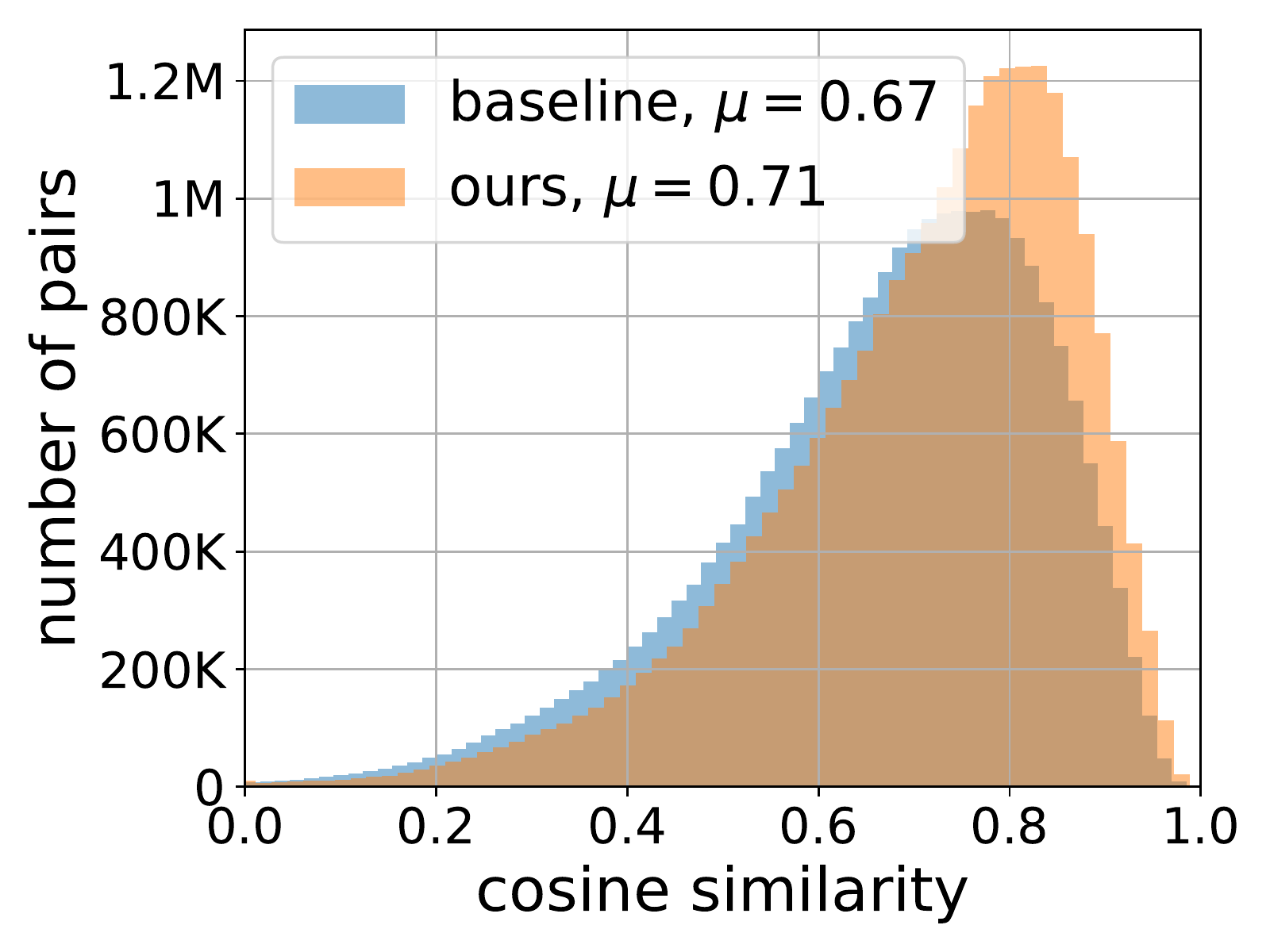}
\end{subfigure}

\begin{subfigure}{.325\linewidth}
  \centering
  \includegraphics[width=0.99\linewidth]{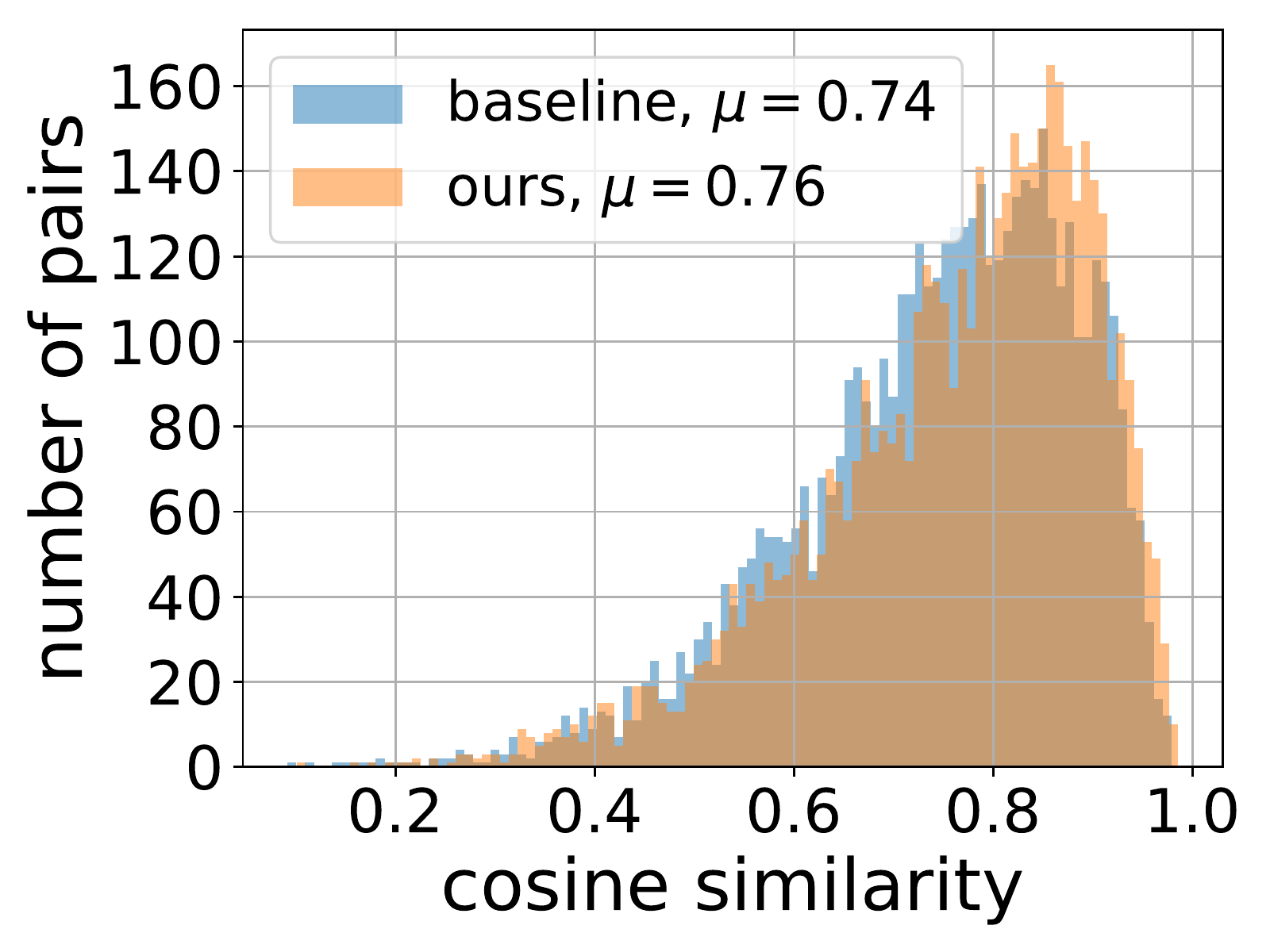}\vspace{-3pt}
  \caption{\textbf{pos pairs: $A_{ii}$}}
\end{subfigure}
\begin{subfigure}{.325\linewidth}
  \centering
  \includegraphics[width=0.99\linewidth]{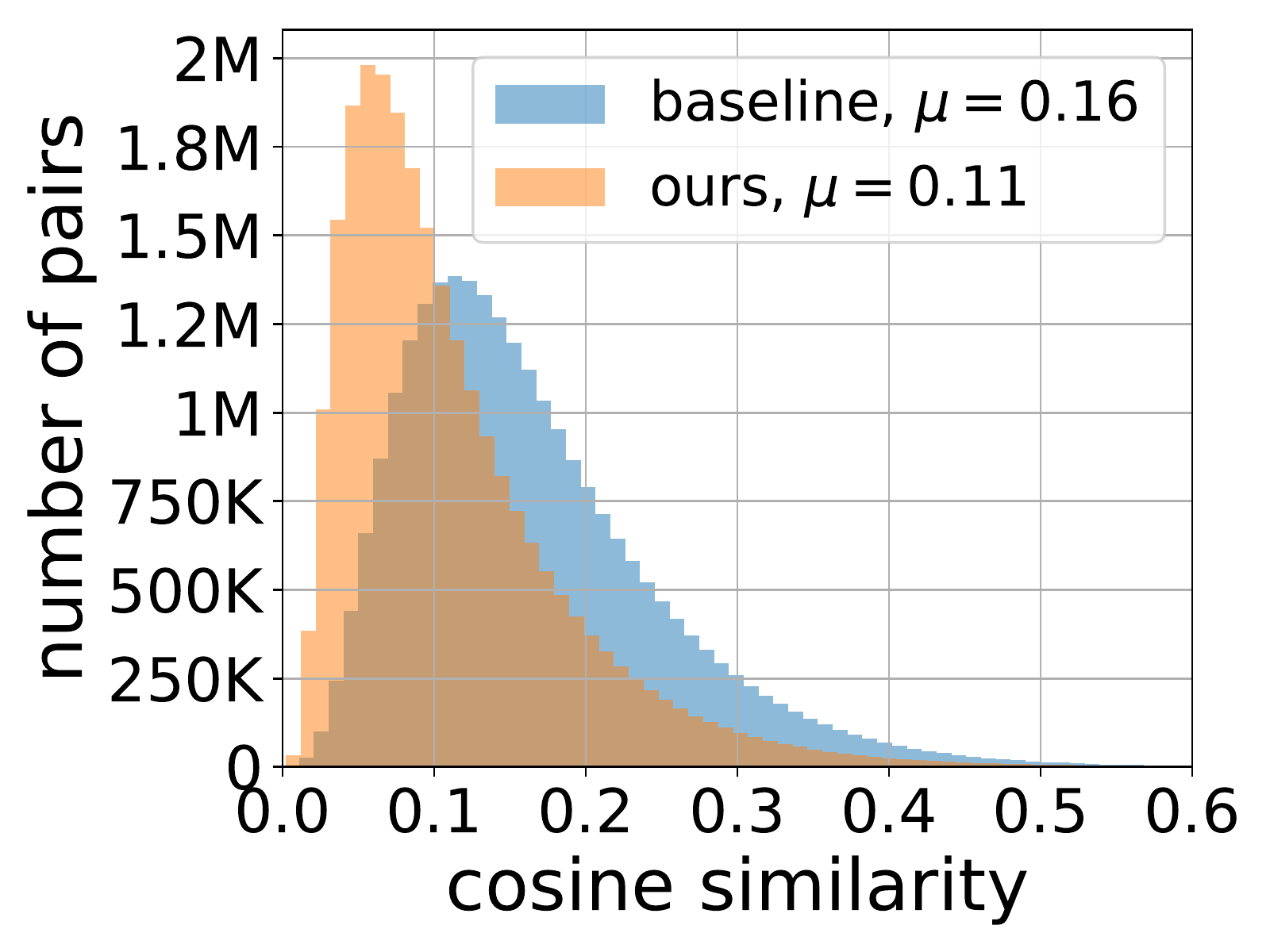}\vspace{-3pt}
  \caption{\textbf{neg pairs: $A_{ij}$}}
\end{subfigure}
\begin{subfigure}{.325\linewidth}
  \centering
  \includegraphics[width=0.99\linewidth]{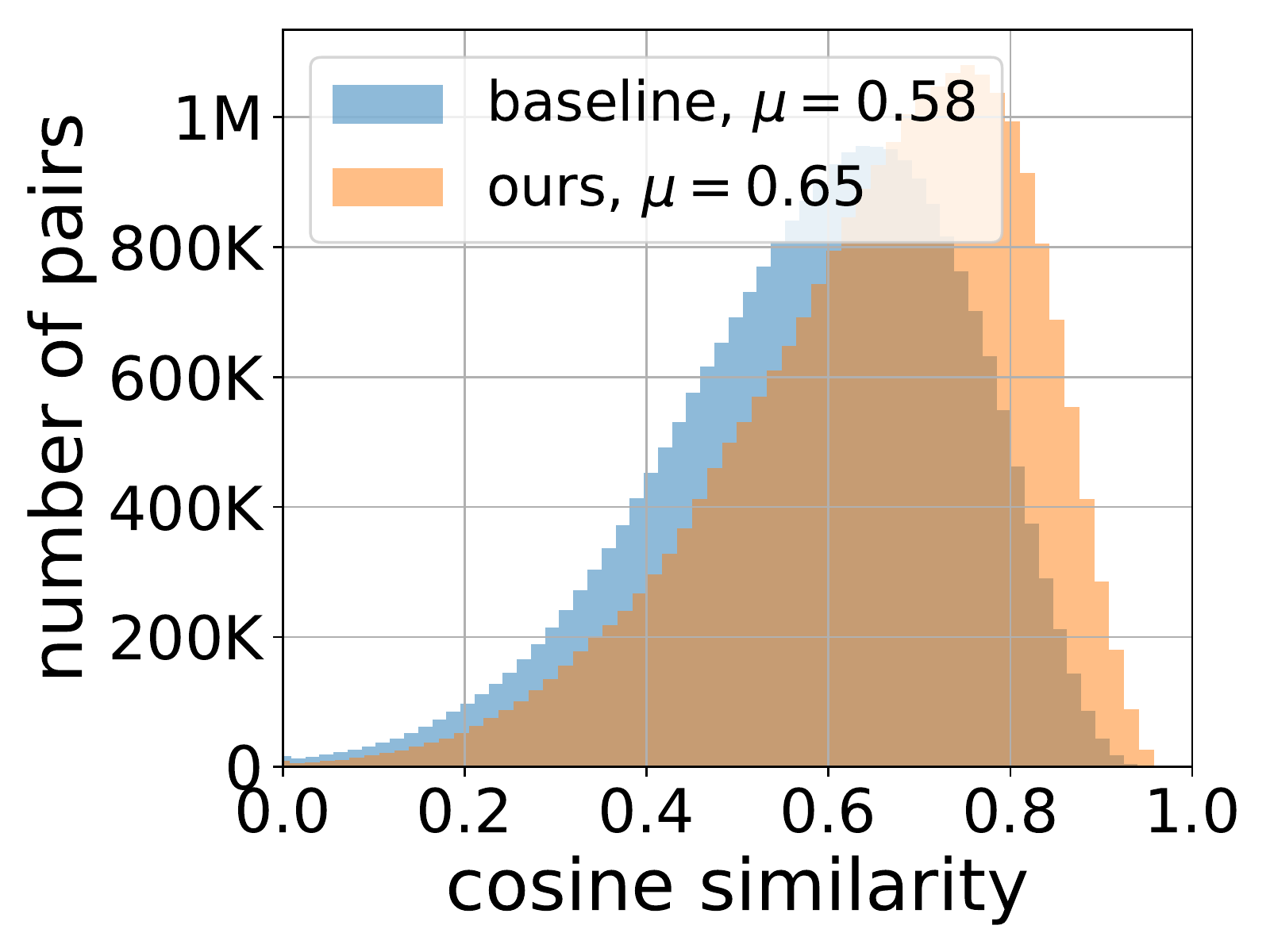}\vspace{-3pt}
  \caption{\textbf{difference: $A^{\Delta}_{ij}$}}
\end{subfigure}\vspace{-3pt}
\caption{CLD has more (dis)similar instances in positive(negative) pairs than baseline MoCo, creating a larger similarity gap.   Columns 1-3 are the \textbf{histograms of cosine similarities} between positive and negative pairs and their differences per the linear projection layer for $f_I(x_i)$ (Row 1) and $f(x_i)$ (Row 2) on ImageNet100.}
\label{fig:CosSim}
\end{figure}
}

\def\figCosSimOrig#1{
\begin{figure}[#1]
\begin{subfigure}{.31\linewidth}
  \centering
  \includegraphics[width=0.99\linewidth]{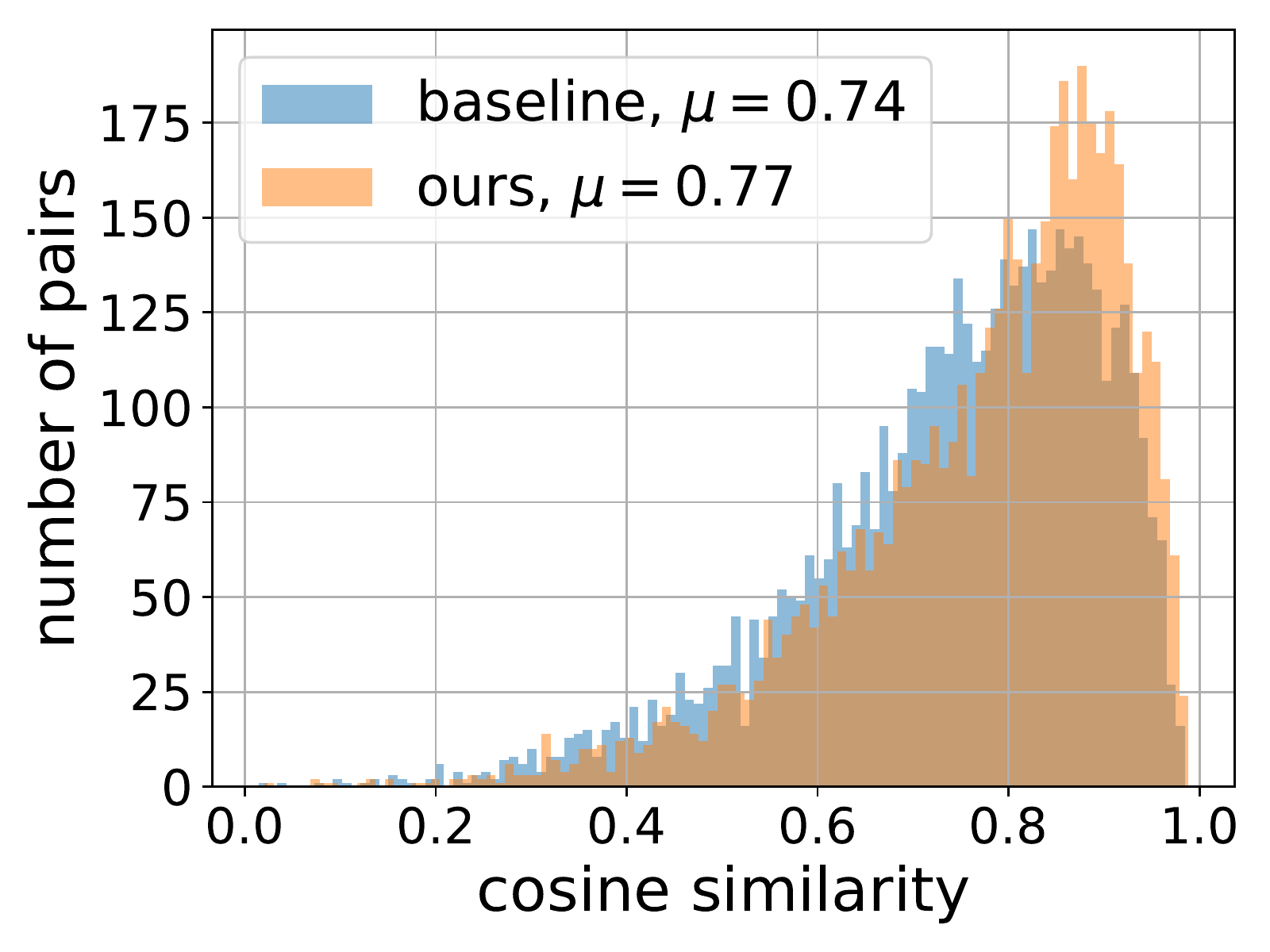}
\end{subfigure}\vspace{0mm}
\begin{subfigure}{.31\linewidth}
  \centering
  \includegraphics[width=0.99\linewidth]{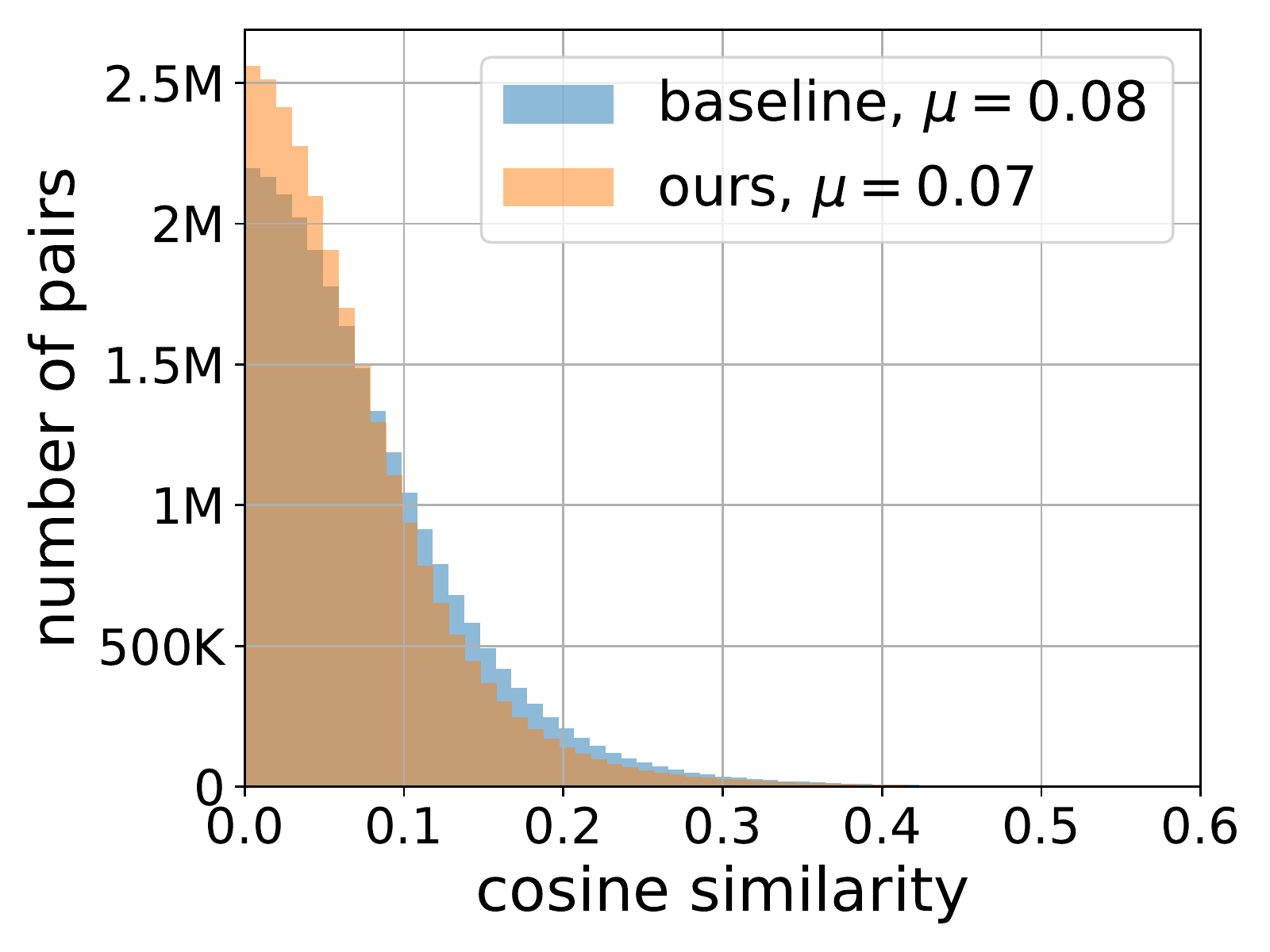}
\end{subfigure}
\begin{subfigure}{.31\linewidth}
  \centering
  \includegraphics[width=0.99\linewidth]{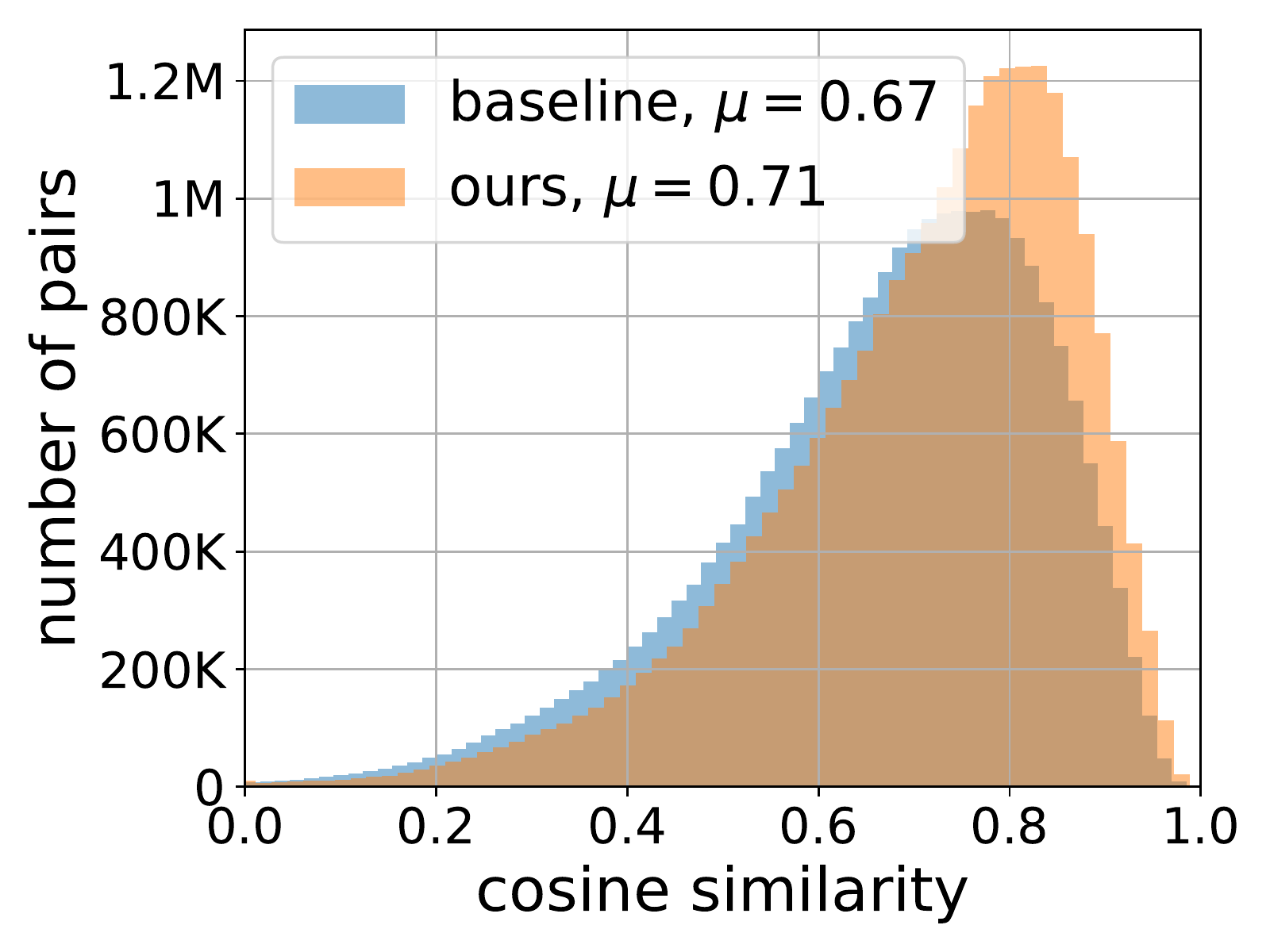}
\end{subfigure}\vspace{-3mm}

\begin{subfigure}{.31\linewidth}
  \centering
  \includegraphics[width=0.99\linewidth]{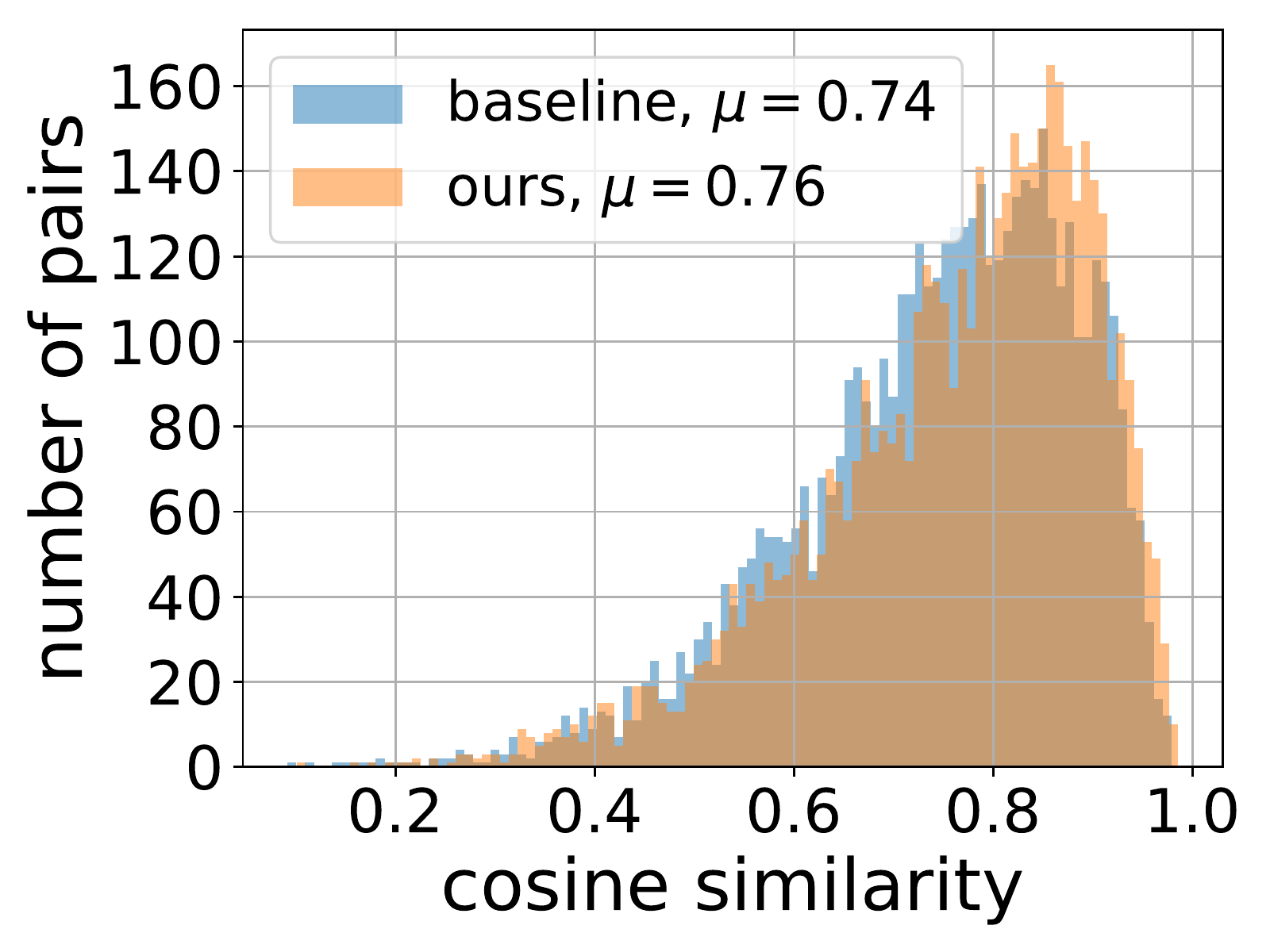}
  \caption{positive pairs: \\$A_{ii}$}
\end{subfigure}\vspace{0mm}
\begin{subfigure}{.31\linewidth}
  \centering
  \includegraphics[width=0.99\linewidth]{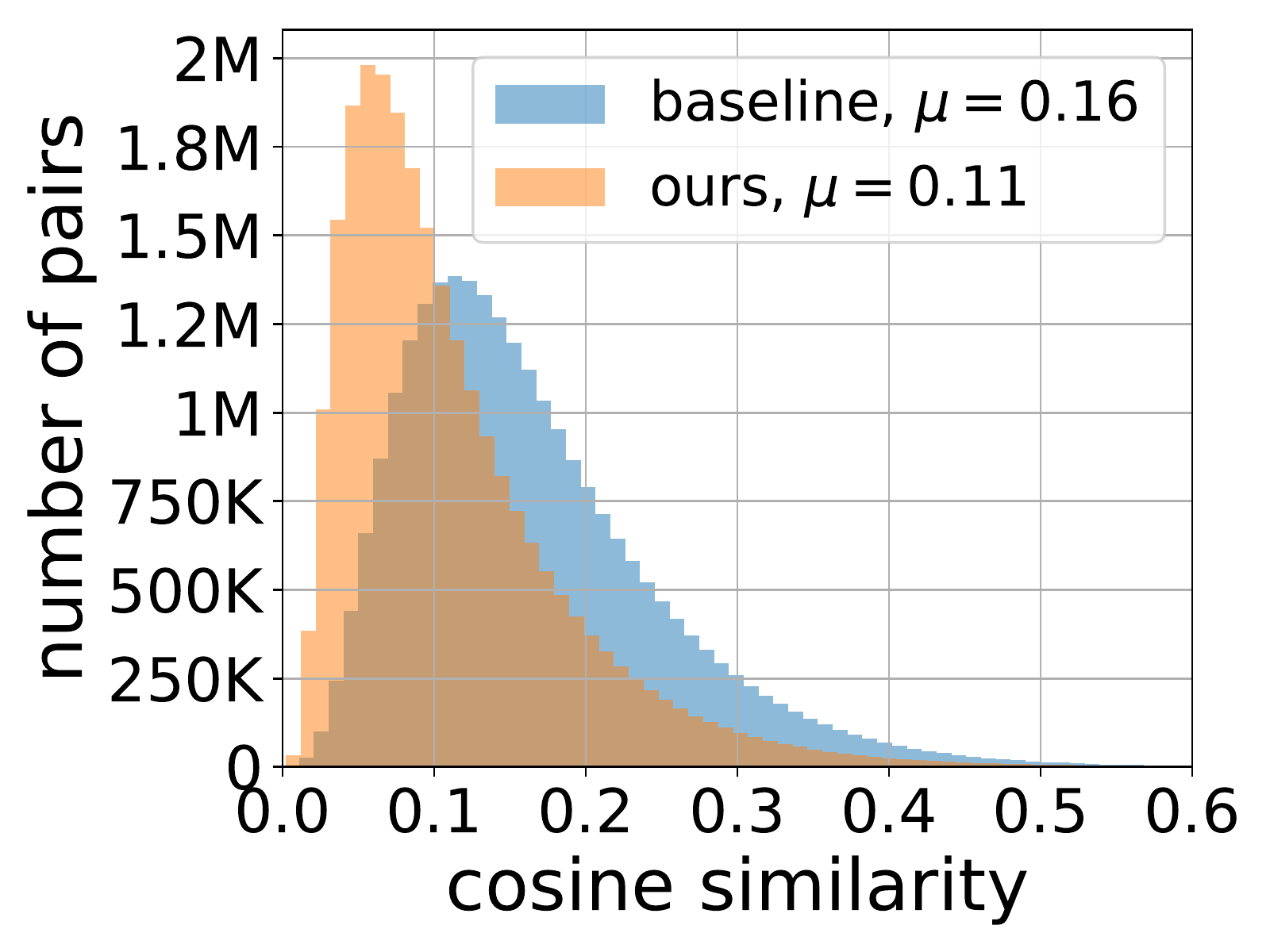}
  \caption{negative pairs: \\$A_{ij}$}
\end{subfigure}
\begin{subfigure}{.31\linewidth}
  \centering
  \includegraphics[width=0.99\linewidth]{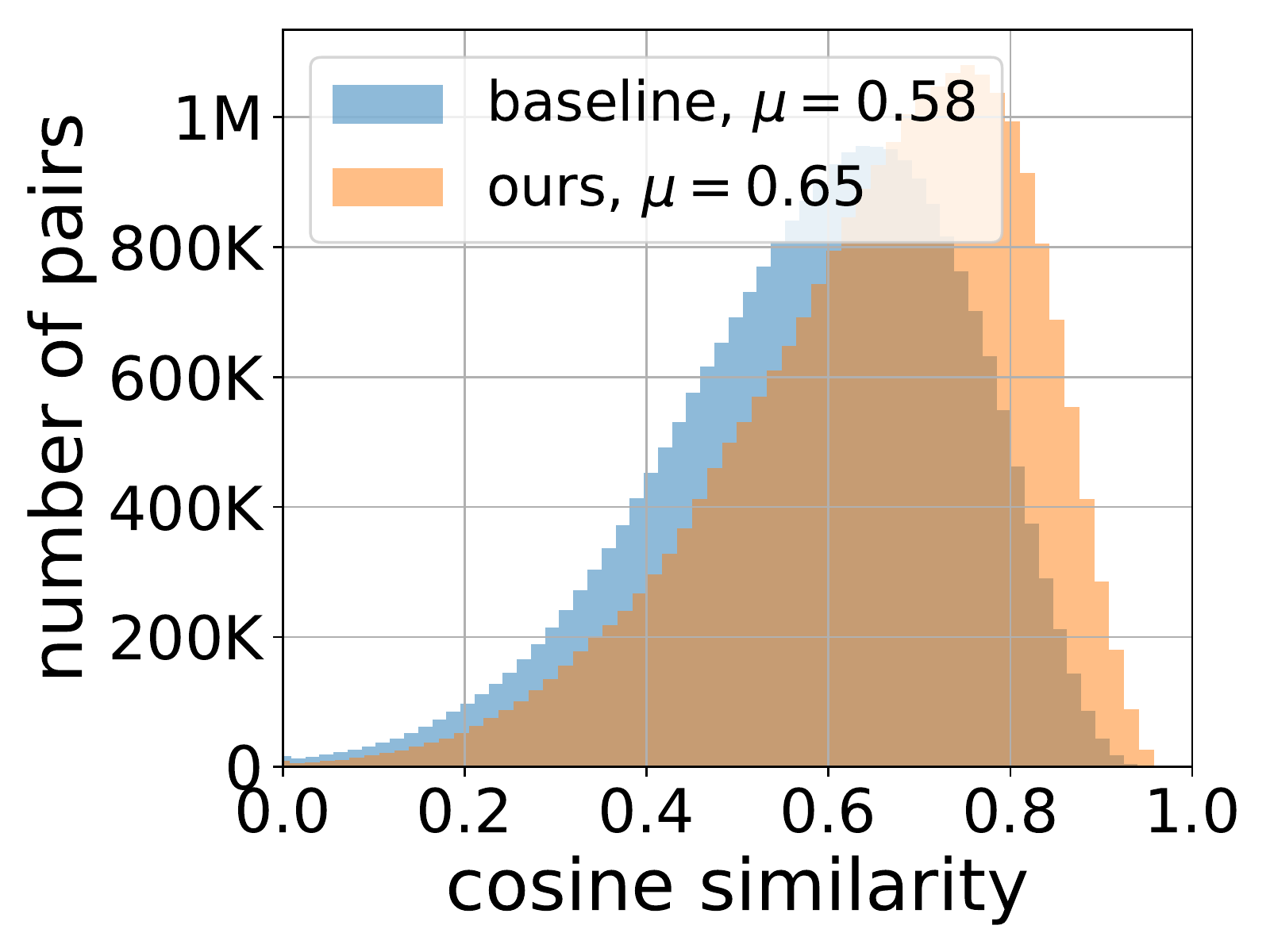}
  \caption{their difference: \\$A^{\Delta}_{ij}$}
\end{subfigure}\vspace{-6pt}
\caption{CLD is better than MoCo at having more (dis)similar instances in positive(negative) pairs, resulting in larger similarity differences between them.   Columns 1-3 are the histograms of cosine similarities between positive and negative pairs and their differences per the linear projection layer for $f_I(x_i)$ (Row 1) and $f(x_i)$ (Row 2) on ImageNet-100 data. }
\label{fig:CosSim}
\end{figure}
}

\def\figTsneLT#1{
\begin{figure}[#1]
\begin{subfigure}{.49\linewidth}
  \centering
  \includegraphics[width=0.95\linewidth]{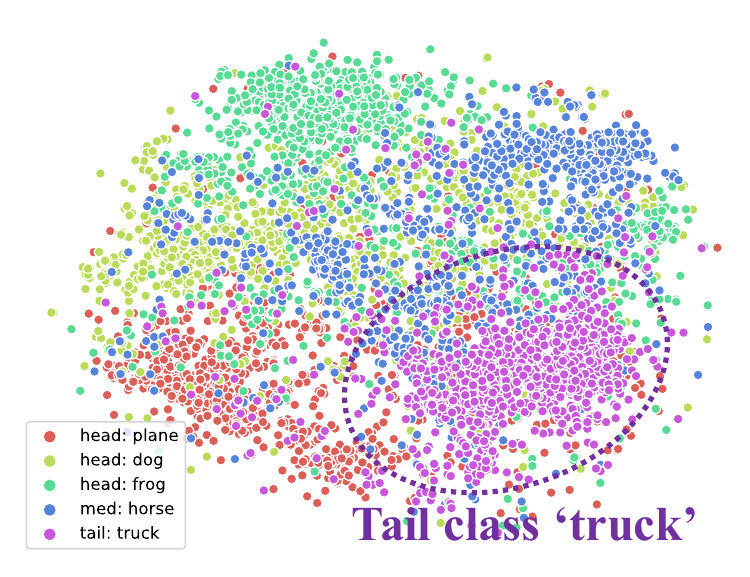}\vspace{-6pt}
  \caption{MoCo}
\end{subfigure}
\begin{subfigure}{.49\linewidth}
  \centering
  \includegraphics[width=0.95\linewidth]{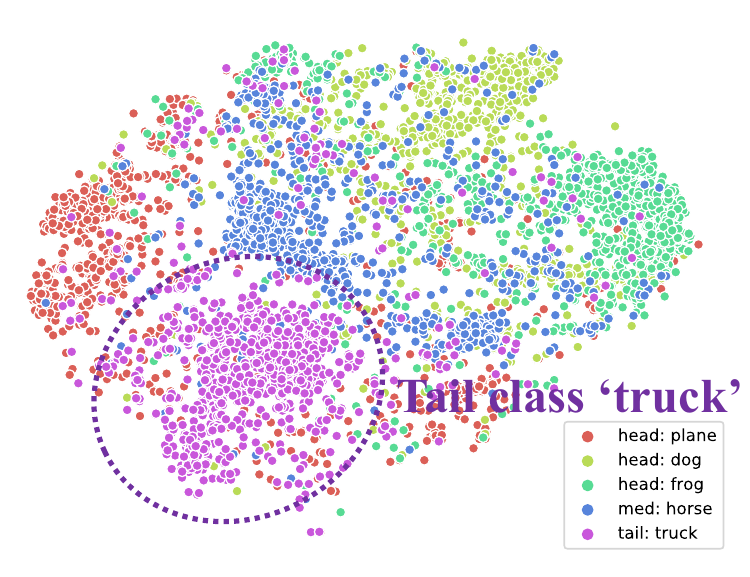}\vspace{-6pt}
  \caption{MoCo+CLD}
\end{subfigure}\vspace{-4pt}
\caption{\textbf{t-SNE feature visualization} of (a) MoCo (b) MoCo + CLD on CIFAR10-LT. Tail class embedding is more compact and better separated from head classes. 
Head and medium-shot classes also have cleaner separation.}
\label{fig:tsne-lt}
\end{figure}
}

\def\figRetrieval#1{
\begin{figure}[#1]
  \centering
  \includegraphics[width=1.0\linewidth]{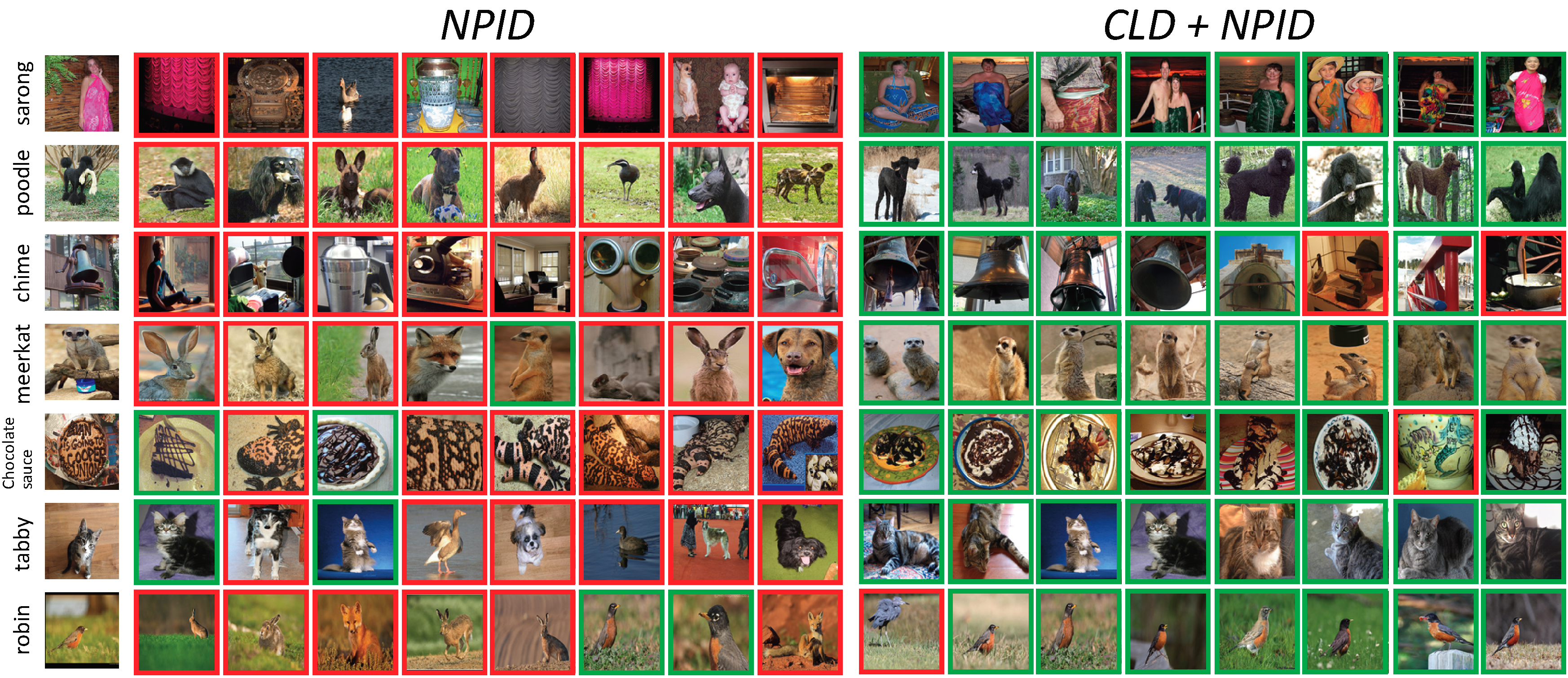}\vspace{-6pt}
\caption{
CLD top retrievals  according to $f_I$ (Columns 10-17) are less distracted by textures than NPID (Columns 2-9) for query images (Column 1) from the ImageNet validation set.
Results are sorted by NPID's performance. Correct retrievals, those in the same category as the query, are outlined in {\color{green} green} and wrong ones in {\color{red} red}.  NPID seems more sensitive to textural appearance (e.g., Rows 1,4,5,7), first retrieve those with similar textures or colors. 
}
\label{fig:Retrieval}
\end{figure}
}

\section{Experiments}

We use ResNet-50 for ImageNet data and ResNet-18 otherwise.
We compare linear classification accuracies on ImageNet, and follow NPID on using kNN accuracies ($k\!=\!200$) for all the small-scale benchmarks. The kNN accuracies are higher and more fitting for metric learning.
Results marked by $\dagger$ are obtained with released code. 

We consider 3 types of datasets.  
{\bf 1) High-correlation}: 
Kitchen-HC is constructed by extracting objects in their bounding boxes from the
 multi-view RGB-D Kitchen dataset \cite{georgakis2016multiview}.  It has 11 categories with highly correlated samples and 20.8K / 4K / 14.4K instances in train / validation / test sets.
{\bf 2) Long-tail}: CIFAR10-LT, CIFAR100-LT and ImageNet-LT \cite{liu2019large}.
{\bf 3) Major benchmarks}: CIFAR \cite{krizhevsky2009learning}, STL10 \cite{coates2011analysis}, ImageNet-100 \cite{tian2019contrastive}, ImageNet \cite{deng2009imagenet}. Following \cite{ye2019unsupervised}, we train models on 5K samples in the {\it train} set and 100K samples in the {\it unlabeled} set, and test on the {\it test} set of STL10.

\subsection{Benchmarking Results}

\figHighCorA{!b}
\figHighCorB{bp}

\noindent\textbf{Results on high-correlation data.}
\label{correlation}
Having highly correlated instances breaks the instance discrimination presumption and causes slow or unstable training.
Accuracies in \fig{HighCorA} and feature visualization in \fig{HighCorB} indeed show that
CLD is much better and fast converging towards a more distinctive feature representation. At Epoch 10, CLD  outperforms by 40\% (23\% vs. 63\%).  CLD outperforms NPID by 9.4\%, when the number of groups used in local clustering is closer to the number of semantic classes in the downstream classification. Likewise, MoCo + CLD outperforms its counterpart MoCo by 5.5\%.

\tabLongtail{!t}
\tabSmallOnly{!h}
\figRetrieval{!t}

\noindent\textbf{Results on long-tailed data.} 
Table \ref{table:long_tail_cifar} shows that
CLD outperforms baselines by a large margin on CIFAR10-LT and CIFAR100-LT.  On ImageNet-LT,
CLD outperforms NPID by 4.5\% per top-5 accuracy, with the largest relative gain (24\%) on few-shot classes;
Our unsupervised CLD even significantly outperforms supervised plain Cross-Entropy (CE) by 8-14\% and is catching up closely with supervised long-tail classifier OLTR (33.3\% vs. 35.6\%).

\tabImageNet{!t}
\noindent\textbf{Results on major benchmarks.}
Table \ref{table:small_datasets} shows that CLD outperforms SOTA on STL10, CIFAR10, CIFAR100 and ImageNet-100. On ImageNet, Table \ref{table:imagenet} shows that CLD consistently outperforms baselines under fair comparison settings: 200 training epochs, standard augmentations \cite{wu2018unsupervised}, and comparable model sizes. 
Adding CLD to InfoMin instead of MoCo produces 7.7\% gain, by using an MLP projection head over feature $f(x)$, a cosine learning scheduler, extra data augmentation \cite{chen2020improved, chen2020simple,tian2020makes}, and a JigSaw branch as in PIRL \cite{misra2019self}.
Fig. \ref{fig:Retrieval} shows CLD retrievals less distracted by textures.

\tabSemi{!t}
\noindent\textbf{Results on semi-supervised learning.} Table \ref{table:semi-super} shows that CLD utilizes annotations far more efficiently, outperforming SOTA (InfoMin) by 6.1\% with only 1\% labeled samples. Baselines and CLDs follow 
OpenSelfSup benchmarks \cite{openselfsup} for fair comparisons.  Baseline results are copied from \cite{openselfsup}.

\tabDetection{!t}
\noindent\textbf{Transfer learning for object detection.}
We test the feature transferability by fine-tuning an ImageNet trained model for Pascal VOC object detection \cite{everingham2010pascal}. Table \ref{table:voc_detection} shows that CLD not only outperforms its supervised learning counterpart by more than 6\%(3\%) in terms of AP in VOC07(VOC07+12), but also surpasses current SOTA of MoCo and MoCo v2.

\subsection{Further Analysis}

\noindent\textbf{Why CLD performs better on long-tailed data?} 
CLD groups similar samples and uses coarse-grained group prototypes instead of instance prototypes.  There are two consequences.
{\bf 1)}
The positive to negative sample ratio is greatly increased from the instance branch to our group branch.  For example,
while each instance is compared against 4,096 negatives (as in MoCo), it is only compared against $k$ negative centroids in our group branch, where $k\le 256$ -- our batch size.  The importance of positives increases from $\frac{1}{4096}$ to $\frac{1}{k}$.   CLD thus achieves better invariant mapping for all the classes, head or tail.   However, the increased ratio is more important for tail classes, as they don't have so many instances to rely on as head classes.  
{\bf 2)}
The imbalance between head and tail classes in the negatives is also reduced in our group branch.  While the distribution of instances in a random mini-batch is long-tailed, it would be more flattened across classes after clustering.  The tail-class negatives would be better represented in the NCE loss.
Fig. \ref{fig:tsne-lt} shows that indeed CLD has clearer class separation than MoCo.

\figTsneLT{t!}
\tabSplitsGroups{!h}

\noindent\textbf{How many groups shall CLD use?}  The ideal number of groups depends on the level of instance correlation, the number of classes, and the batch size.  Table \ref{table:groups} shows that for CIFAR100, CLD
is best when the number of groups is close to the number of classes, although CLD already outperforms MoCo at 10 groups. For ImageNet, the instance correlation is low; since the number of classes of 1,000 is larger than the batch size that our 8 GPUs can afford, we just choose the largest number of groups possible. 
We expect continuous gain with more groups and larger batches afforded by more GPUs.  Nevertheless, our model wins with its merit of the CLD idea instead of a large compute.

\noindent\textbf{Similarity among positives / negatives?} We measure feature (cosine) similarity as $A_{ij}(f) \!=\! <\!\frac{f(x_i)}{\|f(x_i)\|}, \frac{f(x_j')}{\|f(x_j)\|}\!>$, with
$A_{ii}$ ($A_{i,j\neq i}$) for positive (negative) pairs, and their gap is
$A^{\Delta}_{ij}\! =\! A_{ii}\! -\! A_{ij}$.
\fig{CosSim} shows that CLD has higher (lower) similarities between positives (negatives) than MoCo, creating larger gaps of $A^{\Delta}_{ij}$, especially on $f(x_i)$ (\fig{CosSim} Row 2) -- the common feature shared by our instance and group branches, making $f$ a better discriminator than MoCo.  It in turn
 improves $f_I$ (\fig{CosSim} Row 1), the instance branch that runs parallel to the group branch $f_G$.

\figCosSim{!t}

\noindent\textbf{Mutual information characterization?}
We use kNN classification accuracy, Normalized Mutual Information (NMI), and retrieval accuracy $R$ to compare features. 
{\small $\text{NMI}(f,Y) = \frac{I(C|f,Y)}{\sqrt{H(C|f)H(Y)}}$}
reflects global MI between feature $f$ and downstream classification labels $Y$, where $C$ is cluster labels predicted from k-Means clustering of $f$ ($k$ assuming the number of classes), $H(\cdot)$ is entropy, and $I(C|f;Y)$ is the MI between $Y$ and $C$ \cite{strehl2002cluster}. 
The top-1 retrieval accuracy $R(f, Y)$ reflects instance-level mutual information.

Table \ref{table:branch} shows that  $f_I$ is more accurate than  $f_G$ at retrievals and downstream classification.
While $f_G$ has higher NMI, its kNN accuracy is worse than $f_I$.
That is, maximizing global MI would not deliver better downstream classification; maximizing instance-level MI is also important.

\noindent\textbf{Unsupervised hyper-parameter tuning?} 
Unsupervised learning is meant to draw inference from unlabeled data.  However, its hyper-parameters such as our weight $\lambda$ and temperature $T$ are often selected by labeled data in the downstream task.  Self-supervised feature
learning benchmarks pass as a supervised shallow feature learner with a few hyper-parameters.
We explore unsupervised hyper-parameter selection based entirely on the unlabeled data.

\tabUnsupHyper{!t}

We study how the supervised linear accuracy at the downstream can be indicated by unsupervised metrics such as NMI and $R$ between feature $f(x)$ and $f'\!=\!f(x')$.
\fig{unsuper-hyperparam} shows that the linear accuracy is well indicated by $R(f,f')$ for $\lambda$ and by NMI($f,f'$) for temperatures, but neither alone is sufficient.  Their product $\text{NMI}(f, f')\cdot R(f, f')$ turns out to be a promising unsupervised evaluation metric.
\section{Summary}
We extend unsupervised learning to natural data with correlation and long-tail distributions by 
integrating local clustering into contrastive learning.  It discovers between-instance similarity not by direct attraction and repulsion at the instance or group level, but cross-level between instances and groups.  Their batch-wise and cross-view comparisons greatly improve the positive/negative sample ratio for achieving more invariant mapping.   We also propose normalized projection heads and unsupervised hyper-parameter tuning.

Our extensive experimentation and analysis shows that CLD is a lean and powerful add-on to existing SOTA methods, delivering a significant performance boost on all the benchmarks and beating MoCo v2 and SimCLR on every reported performance with a much smaller compute.

\noindent{\bf Acknowledgments.} This work was supported, in part, by Berkeley Deep Drive, US Government Fund through Etegent Technologies on Low-Shot Detection and Semi-supervised Detection, Texas Advanced Computing Center, and NTU NAP and A*STAR via Industry Alignment Fund.

{\small
\bibliographystyle{ieee_fullname}
\bibliography{egbib}

\begin{thebibliography}{10}\itemsep=-1pt

\bibitem{bachman2019amdim}
Philip Bachman, R~Devon Hjelm, and William Buchwalter.
\newblock Learning representations by maximizing mutual information across
  views.
\newblock In {\em NeurIPS}, 2019.

\bibitem{bernardis2010finding}
Elena Bernardis and Stella~X. Yu.
\newblock Finding dots: Segmentation as popping out regions from boundaries.
\newblock In {\em CVPR}, 2010.

\bibitem{buchta2012spherical}
Christian Buchta, Martin Kober, Ingo Feinerer, and Kurt Hornik.
\newblock Spherical k-means clustering.
\newblock {\em Journal of Statistical Software}, 2012.

\bibitem{caron2018deep}
Mathilde Caron, Piotr Bojanowski, Armand Joulin, and Matthijs Douze.
\newblock Deep clustering for unsupervised learning of visual features.
\newblock In {\em ECCV}, 2018.

\bibitem{caron2019unsupervised}
Mathilde Caron, Piotr Bojanowski, Julien Mairal, and Armand Joulin.
\newblock Unsupervised pre-training of image features on non-curated data.
\newblock In {\em ICCV}, 2019.

\bibitem{caron2020unsupervised}
Mathilde Caron, Ishan Misra, Julien Mairal, Priya Goyal, Piotr Bojanowski, and
  Armand Joulin.
\newblock Unsupervised learning of visual features by contrasting cluster
  assignments.
\newblock {\em Advances in Neural Information Processing Systems}, 33, 2020.

\bibitem{chen2020simple}
Ting Chen, Simon Kornblith, Mohammad Norouzi, and Geoffrey Hinton.
\newblock A simple framework for contrastive learning of visual
  representations.
\newblock {\em {\it{arXiv preprint arXiv:2002.05709}}}, 2020.

\bibitem{chen2020improved}
Xinlei Chen, Haoqi Fan, Ross Girshick, and Kaiming He.
\newblock Improved baselines with momentum contrastive learning.
\newblock {\em arXiv preprint arXiv:2003.04297}, 2020.

\bibitem{chen2020exploring}
Xinlei Chen and Kaiming He.
\newblock Exploring simple siamese representation learning.
\newblock {\em arXiv preprint arXiv:2011.10566}, 2020.

\bibitem{coates2011analysis}
Adam Coates, Andrew Ng, and Honglak Lee.
\newblock An analysis of single-layer networks in unsupervised feature
  learning.
\newblock In {\em AISTATS}, 2011.

\bibitem{de2006discriminative}
Fernando De~la Torre and Takeo Kanade.
\newblock Discriminative cluster analysis.
\newblock In {\em ICML}, 2006.

\bibitem{dempster1977maximum}
Arthur~P Dempster, Nan~M Laird, and Donald~B Rubin.
\newblock Maximum likelihood from incomplete data via the em algorithm.
\newblock {\em Journal of the Royal Statistical Society: Series B
  (Methodological)}, 1977.

\bibitem{deng2009imagenet}
Jia Deng, Wei Dong, Richard Socher, Li-Jia Li, Kai Li, and Li Fei-Fei.
\newblock Imagenet: A large-scale hierarchical image database.
\newblock In {\em CVPR}, pages 248--255, 2009.

\bibitem{ding2007adaptive}
Chris Ding and Tao Li.
\newblock Adaptive dimension reduction using discriminant analysis and k-means
  clustering.
\newblock In {\em ICML}, 2007.

\bibitem{doersch2015unsupervised}
Carl Doersch, Abhinav Gupta, and Alexei~A Efros.
\newblock Unsupervised visual representation learning by context prediction.
\newblock In {\em ICCV}, 2015.

\bibitem{donahue2017adversarial}
Jeff Donahue, Philipp Kr{\"a}henb{\"u}hl, and Trevor Darrell.
\newblock Adversarial feature learning.
\newblock In {\em ICLR}, 2017.

\bibitem{dosovitskiy2015discriminative}
Alexey Dosovitskiy, Philipp Fischer, Jost~Tobias Springenberg, Martin
  Riedmiller, and Thomas Brox.
\newblock Discriminative unsupervised feature learning with exemplar
  convolutional neural networks.
\newblock {\em TPAMI}, 2015.

\bibitem{everingham2010pascal}
Mark Everingham, Luc Van~Gool, Christopher~KI Williams, John Winn, and Andrew
  Zisserman.
\newblock The pascal visual object classes (voc) challenge.
\newblock {\em IJCV}, 88(2):303--338, 2010.

\bibitem{gan2007data}
Guojun Gan, Chaoqun Ma, and Jianhong Wu.
\newblock {\em Data clustering: theory, algorithms, and applications}.
\newblock SIAM, 2007.

\bibitem{georgakis2016multiview}
Georgios Georgakis, Md~Alimoor Reza, Arsalan Mousavian, Phi-Hung Le, and Jana
  Ko{\v{s}}eck{\'a}.
\newblock Multiview rgb-d dataset for object instance detection.
\newblock In {\em 3DV}, 2016.

\bibitem{gidaris2018unsupervised}
Spyros Gidaris, Praveer Singh, Nikos Komodakis, et~al.
\newblock Unsupervised representation learning by predicting image rotations.
\newblock In {\em ICLR}, 2018.

\bibitem{goyal2019scaling}
Priya Goyal, Dhruv Mahajan, Abhinav Gupta, and Ishan Misra.
\newblock Scaling and benchmarking self-supervised visual representation
  learning.
\newblock {\em {\it{arXiv preprint arXiv:1905.01235}}}, 2019.

\bibitem{grill2020bootstrap}
Jean-Bastien Grill, Florian Strub, Florent Altch{\'e}, Corentin Tallec,
  Pierre~H Richemond, Elena Buchatskaya, Carl Doersch, Bernardo~Avila Pires,
  Zhaohan~Daniel Guo, Mohammad~Gheshlaghi Azar, et~al.
\newblock Bootstrap your own latent: A new approach to self-supervised
  learning.
\newblock {\em arXiv preprint arXiv:2006.07733}, 2020.

\bibitem{gutmann2012noise}
Michael~U Gutmann and Aapo Hyv{\"a}rinen.
\newblock Noise-contrastive estimation of unnormalized statistical models, with
  applications to natural image statistics.
\newblock {\em Journal of Machine Learning Research}, 2012.

\bibitem{hadsell2006dimensionality}
Raia Hadsell, Sumit Chopra, and Yann LeCun.
\newblock Dimensionality reduction by learning an invariant mapping.
\newblock In {\em CVPR}, 2006.

\bibitem{he2020momentum}
Kaiming He, Haoqi Fan, Yuxin Wu, Saining Xie, and Ross Girshick.
\newblock Momentum contrast for unsupervised visual representation learning.
\newblock In {\em CVPR}, 2020.

\bibitem{he2017mask}
Kaiming He, Georgia Gkioxari, Piotr Doll{\'a}r, and Ross Girshick.
\newblock Mask r-cnn.
\newblock In {\em ICCV}, 2017.

\bibitem{he2016deep}
Kaiming He, Xiangyu Zhang, Shaoqing Ren, and Jian Sun.
\newblock Deep residual learning for image recognition.
\newblock In {\em CVPR}, 2016.

\bibitem{henaff2019data}
Olivier~J H{\'e}naff, Aravind Srinivas, Jeffrey De~Fauw, Ali Razavi, Carl
  Doersch, SM Eslami, and Aaron van~den Oord.
\newblock Data-efficient image recognition with contrastive predictive coding.
\newblock {\em arXiv preprint arXiv:1905.09272}, 2019.

\bibitem{hjelm2018learning}
R~Devon Hjelm, Alex Fedorov, Samuel Lavoie-Marchildon, Karan Grewal, Phil
  Bachman, Adam Trischler, and Yoshua Bengio.
\newblock Learning deep representations by mutual information estimation and
  maximization.
\newblock {\em arXiv preprint arXiv:1808.06670}, 2018.

\bibitem{huang2017densely}
Gao Huang, Zhuang Liu, Laurens Van Der~Maaten, and Kilian~Q Weinberger.
\newblock Densely connected convolutional networks.
\newblock In {\em CVPR}, 2017.

\bibitem{hwang2019segsort}
Jyh-Jing Hwang, Stella~X. Yu, Jianbo Shi, Maxwell~D Collins, Tien-Ju Yang, Xiao
  Zhang, and Liang-Chieh Chen.
\newblock Segsort: Segmentation by discriminative sorting of segments.
\newblock In {\em ICCV}, 2019.

\bibitem{jenni2018self}
Simon Jenni and Paolo Favaro.
\newblock Self-supervised feature learning by learning to spot artifacts.
\newblock In {\em CVPR}, 2018.

\bibitem{ji19invariant}
Xu Ji, Jo{\~{a}}o~F. Henriques, and Andrea Vedaldi.
\newblock Invariant information clustering for unsupervised image
  classification and segmentation.
\newblock In {\em ICCV}, 2019.

\bibitem{kanungo2002efficient}
Tapas Kanungo, David~M Mount, Nathan~S Netanyahu, Christine~D Piatko, Ruth
  Silverman, and Angela~Y Wu.
\newblock An efficient k-means clustering algorithm: Analysis and
  implementation.
\newblock {\em IEEE transactions on pattern analysis and machine intelligence},
  2002.

\bibitem{krizhevsky2009learning}
Alex Krizhevsky, Geoffrey Hinton, et~al.
\newblock Learning multiple layers of features from tiny images.
\newblock {\em Citeseer}, 2009.

\bibitem{krizhevsky2012imagenet}
Alex Krizhevsky, Ilya Sutskever, and Geoffrey~E Hinton.
\newblock Imagenet classification with deep convolutional neural networks.
\newblock In {\em NeurIPS}, 2012.

\bibitem{larsson2017colorization}
Gustav Larsson, Michael Maire, and Gregory Shakhnarovich.
\newblock Colorization as a proxy task for visual understanding.
\newblock In {\em CVPR}, 2017.

\bibitem{li2020prototypical}
Junnan Li, Pan Zhou, Caiming Xiong, Richard Socher, and Steven~CH Hoi.
\newblock Prototypical contrastive learning of unsupervised representations.
\newblock {\em arXiv preprint arXiv:2005.04966}, 2020.

\bibitem{liu2019large}
Ziwei Liu, Zhongqi Miao, Xiaohang Zhan, Jiayun Wang, Boqing Gong, and Stella~X
  Yu.
\newblock Large-scale long-tailed recognition in an open world.
\newblock In {\em CVPR}, 2019.

\bibitem{maire2011object}
Michael Maire, Stella~X. Yu, and Pietro Perona.
\newblock Object detection and segmentation from joint embedding of parts and
  pixels.
\newblock In {\em ICCV}, 2011.

\bibitem{misra2019self}
Ishan Misra and Laurens van~der Maaten.
\newblock Self-supervised learning of pretext-invariant representations.
\newblock {\em arXiv preprint arXiv:1912.01991}, 2019.

\bibitem{nie2011spectral}
Feiping Nie, Zinan Zeng, Ivor~W Tsang, Dong Xu, and Changshui Zhang.
\newblock Spectral embedded clustering: A framework for in-sample and
  out-of-sample spectral clustering.
\newblock {\em IEEE Transactions on Neural Networks}, 2011.

\bibitem{noroozi2016unsupervised}
Mehdi Noroozi and Paolo Favaro.
\newblock Unsupervised learning of visual representations by solving jigsaw
  puzzles.
\newblock In {\em ECCV}, 2016.

\bibitem{oord2018representation}
Aaron van~den Oord, Yazhe Li, and Oriol Vinyals.
\newblock Representation learning with contrastive predictive coding.
\newblock {\em arXiv preprint arXiv:1807.03748}, 2018.

\bibitem{paninski2003estimation}
Liam Paninski.
\newblock Estimation of entropy and mutual information.
\newblock {\em Neural computation}, 2003.

\bibitem{pathak2016context}
Deepak Pathak, Philipp Krahenbuhl, Jeff Donahue, Trevor Darrell, and Alexei~A
  Efros.
\newblock Context encoders: Feature learning by inpainting.
\newblock In {\em CVPR}, 2016.

\bibitem{peng2018megdet}
Chao Peng, Tete Xiao, Zeming Li, Yuning Jiang, Xiangyu Zhang, Kai Jia, Gang Yu,
  and Jian Sun.
\newblock Megdet: A large mini-batch object detector.
\newblock In {\em CVPR}, 2018.

\bibitem{poole2019variational}
Ben Poole, Sherjil Ozair, Aaron van~den Oord, Alexander~A Alemi, and George
  Tucker.
\newblock On variational bounds of mutual information.
\newblock {\em arXiv preprint arXiv:1905.06922}, 2019.

\bibitem{shi2000normalized}
Jianbo Shi and Jitendra Malik.
\newblock Normalized cuts and image segmentation.
\newblock {\em IEEE Transactions on pattern analysis and machine intelligence},
  2000.

\bibitem{strehl2002cluster}
Alexander Strehl and Joydeep Ghosh.
\newblock Cluster ensembles---a knowledge reuse framework for combining
  multiple partitions.
\newblock {\em Journal of machine learning research}, 2002.

\bibitem{tian2014learning}
Fei Tian, Bin Gao, Qing Cui, Enhong Chen, and Tie-Yan Liu.
\newblock Learning deep representations for graph clustering.
\newblock In {\em AAAI}, 2014.

\bibitem{tian2019contrastive}
Yonglong Tian, Dilip Krishnan, and Phillip Isola.
\newblock Contrastive multiview coding.
\newblock {\em {\it{arXiv preprint arXiv:1906.05849}}}, 2019.

\bibitem{tian2020makes}
Yonglong Tian, Chen Sun, Ben Poole, Dilip Krishnan, Cordelia Schmid, and
  Phillip Isola.
\newblock What makes for good views for contrastive learning.
\newblock {\em arXiv preprint arXiv:2005.10243}, 2020.

\bibitem{tschannen2019mutual}
Michael Tschannen, Josip Djolonga, Paul~K Rubenstein, Sylvain Gelly, and Mario
  Lucic.
\newblock On mutual information maximization for representation learning.
\newblock {\em arXiv preprint arXiv:1907.13625}, 2019.

\bibitem{van2009learning}
Laurens Van Der~Maaten.
\newblock Learning a parametric embedding by preserving local structure.
\newblock In {\em Artificial Intelligence and Statistics}, 2009.

\bibitem{von2007tutorial}
Ulrike Von~Luxburg.
\newblock A tutorial on spectral clustering.
\newblock {\em Statistics and computing}, 2007.

\bibitem{wu2018unsupervised}
Zhirong Wu, Yuanjun Xiong, Stella~X Yu, and Dahua Lin.
\newblock Unsupervised feature learning via non-parametric instance
  discrimination.
\newblock In {\em CVPR}, 2018.

\bibitem{xie2016unsupervised}
Junyuan Xie, Ross Girshick, and Ali Farhadi.
\newblock Unsupervised deep embedding for clustering analysis.
\newblock In {\em ICML}, 2016.

\bibitem{yang2016joint}
Jianwei Yang, Devi Parikh, and Dhruv Batra.
\newblock Joint unsupervised learning of deep representations and image
  clusters.
\newblock In {\em CVPR}, 2016.

\bibitem{yang2010image}
Yi Yang, Dong Xu, Feiping Nie, Shuicheng Yan, and Yueting Zhuang.
\newblock Image clustering using local discriminant models and global
  integration.
\newblock {\em TIP}, 2010.

\bibitem{ye2008discriminative}
Jieping Ye, Zheng Zhao, and Mingrui Wu.
\newblock Discriminative k-means for clustering.
\newblock In {\em NIPs}, 2008.

\bibitem{ye2019unsupervised}
Mang Ye, Xu Zhang, Pong~C Yuen, and Shih-Fu Chang.
\newblock Unsupervised embedding learning via invariant and spreading instance
  feature.
\newblock In {\em CVPR}, 2019.

\bibitem{yu2001segmentation}
Stella~X. Yu and Jianbo Shi.
\newblock Segmentation with pairwise attraction and repulsion.
\newblock In {\em ICCV}, 2001.

\bibitem{yu2001understanding}
Stella~X. Yu and Jianbo Shi.
\newblock Understanding popout through repulsion.
\newblock In {\em CVPR}, 2001.

\bibitem{stella2003multiclass}
Stella~X. Yu and Jianbo Shi.
\newblock Multiclass spectral clustering.
\newblock In {\em CVPR}, 2003.

\bibitem{zhai2019s4l}
Xiaohua Zhai, Avital Oliver, Alexander Kolesnikov, and Lucas Beyer.
\newblock S4l: Self-supervised semi-supervised learning.
\newblock In {\em Proceedings of the IEEE international conference on computer
  vision}, pages 1476--1485, 2019.

\bibitem{zhan2019self}
Xiaohang Zhan, Xingang Pan, Ziwei Liu, Dahua Lin, and Chen~Change Loy.
\newblock Self-supervised learning via conditional motion propagation.
\newblock In {\em CVPR}, 2019.

\bibitem{openselfsup}
Xiaohang Zhan, Jiahao Xie, Ziwei Liu, Dahua Lin, and Chen Change~Loy.
\newblock {OpenSelfSup}: Open mmlab self-supervised learning toolbox and
  benchmark.
\newblock 2020.

\bibitem{zhang2016colorful}
Richard Zhang, Phillip Isola, and Alexei~A Efros.
\newblock Colorful image colorization.
\newblock In {\em ECCV}, 2016.

\bibitem{zhuang2019local}
Chengxu Zhuang, Alex~Lin Zhai, Daniel Yamins, , et~al.
\newblock Local aggregation for unsupervised learning of visual embeddings.
\newblock In {\em ICCV}, 2019.

\end{thebibliography}
}

\clearpage

\def\figKitchen#1{
\begin{figure}[#1]
  \centering
  \includegraphics[width=.99\linewidth]{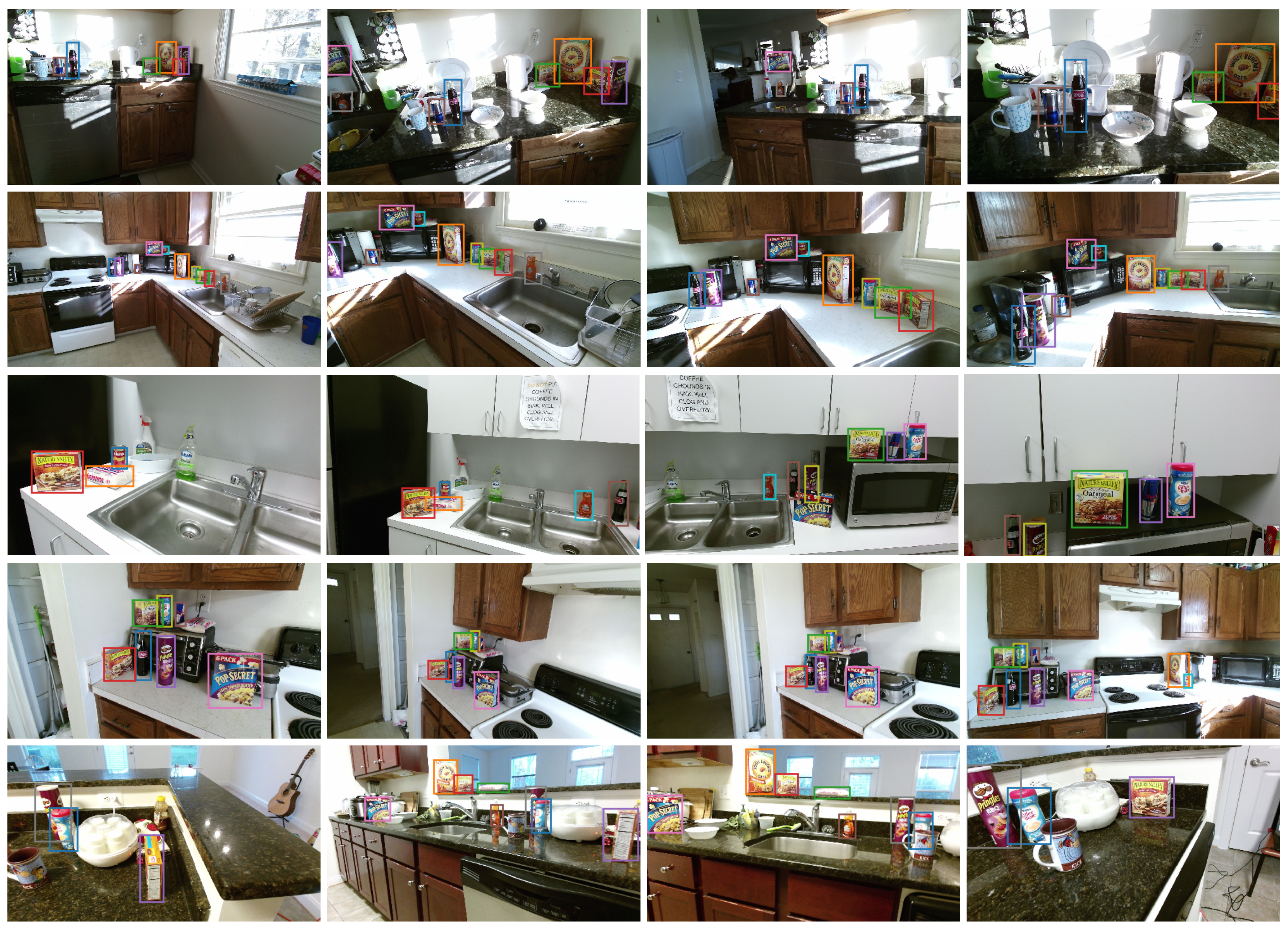}
\vspace{-2mm}
\caption{Samples of multi-view RGB-D dataset Kitchen \cite{georgakis2016multiview}. Instances of the same category captured from different perspectives are highly correlated. The high-correlation dataset Kitchen-HC is constructed from Kitchen by extracting objects in their bounding boxes.}
\label{fig:kitchen}
\end{figure}
}

\def\figinfoNCEK#1{
\begin{figure}[#1]
\begin{subfigure}{.48\textwidth}
  \centering
  \includegraphics[width=.99\linewidth]{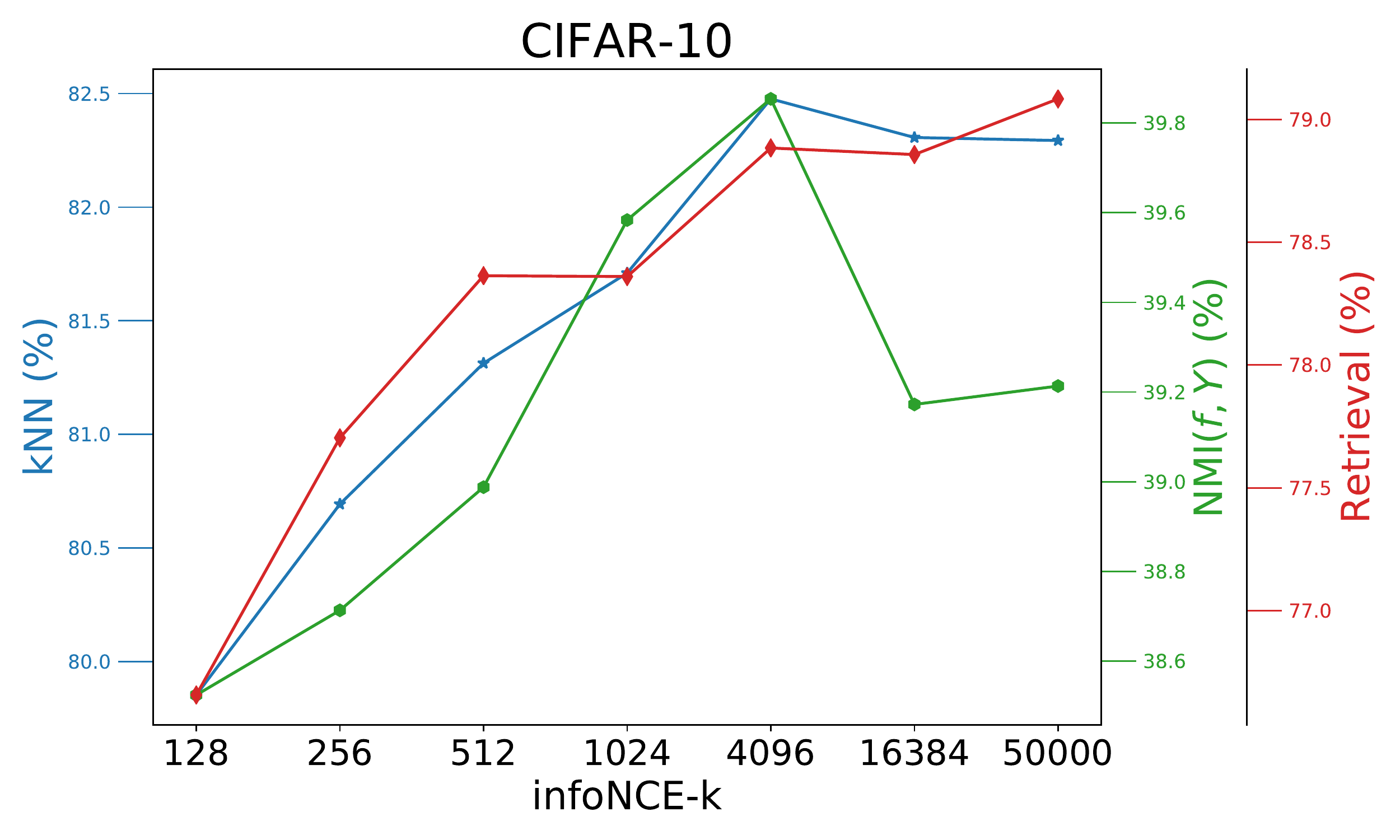}
\end{subfigure}
\begin{subfigure}{.48\textwidth}
  \centering
  \includegraphics[width=.99\linewidth]{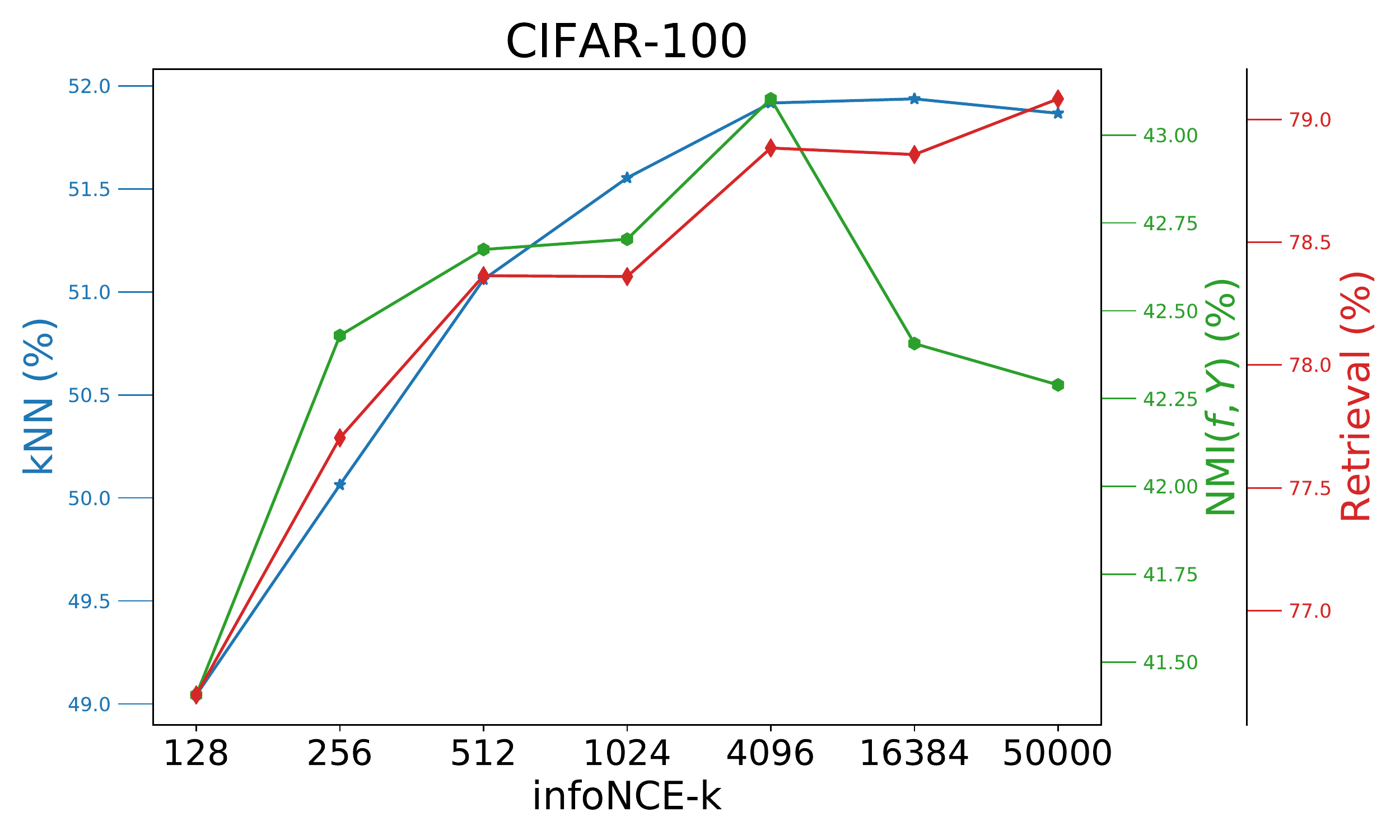}
\end{subfigure}\vspace{-4mm}
\caption{MoCo trained with different memory bank sizes are evaluated with NMI, retrieval and kNN accuracy. 
While a larger memory bank improves the retrieval performance,
 the classification accuracy and NMI score do not always increase:
the NMI score drops sharply due to a large negative/positive ratio. 
There is a trade-off for best performance at downstream classification.
}
\label{fig:infoNCEK}
\end{figure}
}

\def\figRetrieval#1{
\begin{figure}[#1]
  \centering
  \includegraphics[width=1.0\linewidth]{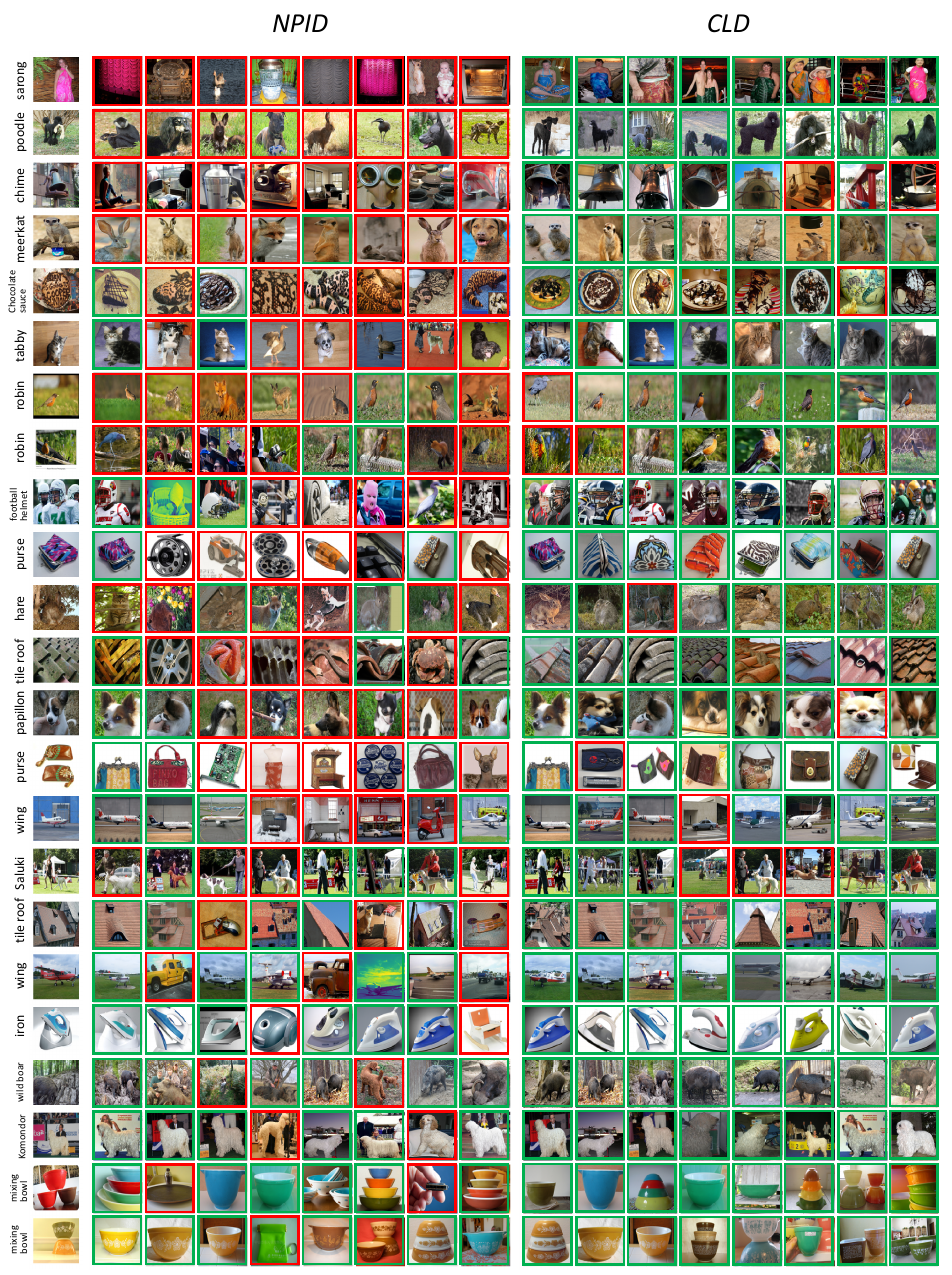}\vspace{-8pt}
\caption{Comparisons of top {{\bf retrieves}} by NPID (Columns 2-9) and NPID+CLD (Columns 10-17) according to $f_I$ for the query images (Column 1) from the ImageNet validation set.  The results are sorted by NPID's performance: Retrievals with the same category as the query are outlined in {\color{green} green} and otherwise in {\color{red} red}. NPID seems to be much more sensitive to textural appearance (e.g., Rows 1,4,5,7), first retrieve those with similar textures or colors.  Integrated with CLD, NPID+CLD is able to retrieve semantically similar samples. (Zoom in for details)}
\label{fig:Retrieval}
\end{figure}
}

\def\tabClust#1{
\begin{table}[#1]
\tablestyle{15.0pt}{1.0}
\begin{tabular}{c|cc}
\shline
group & spectral & k-Means \\
\shline
10 & 77.1\% & 78.9\% \\
64 & 74.5\% & 76.3\% \\
128 & 72.6\% & 73.4\% \\
256 & 70.5\% & 70.8\% \\
\shline
\end{tabular}
\caption{Top-1 kNN accuracies on Kitchen-HC under different group numbers for different clustering methods.}
\label{table:cluster}
\end{table}
}

\def\tabComponents#1{
\begin{table}[#1]
\tablestyle{2.0pt}{1.0}
\begin{tabular}{ccc|cc}\\
\shline
NPID+CLD & Subspace & Cross-augmentation & CIFAR-10 & CIFAR-100\\
\shline
\xmark &  &  & 80.8\% & 51.6\% \\
\cmark & share &  & 82.7\% & 53.3\% \\
\cmark & separate &  & 84.2\% & 55.0\% \\
\cmark & separate & \cmark & 86.5\% & 57.5\% \\
\shline
\end{tabular}
\caption{Ablation study on various components of our method, i.e. adding the cross-level discrimination, projecting the representation to two different spaces, and using cross-augmentation comparison between $x_i$ and $x_i'$. kNN top-1 accuracy is reported here. }
\label{table:components}
\end{table}
}

\def\tabLambda#1{
\begin{table}[#1]
\tablestyle{6.0pt}{1.0}
\begin{tabular}{l||c|c||c|c}
\shline
\multirow{3}{*}{} & \multicolumn{2}{c||}{NPID+CLD} & \multicolumn{2}{c}{MoCo+CLD} \\
\cline{2-5}
 & top-1 (\%) & top-5 (\%) & top-1 (\%) & top-5 (\%)\\ [.1em]
\hline
$\lambda=0$ & 75.3 & 92.4 & 77.6 & 93.8\\ [.1em]
\hline
$\lambda=0.1$ & 78.8 & 94.4 & 80.3 & 95.0\\ [.1em]
$\lambda=0.25$ & \bf{79.7} & \bf{95.1} & \bf{81.7} & \bf{95.7}\\ [.1em]
$\lambda=0.50$ & 78.9 & 94.4 & 80.5 & 95.2\\ [.1em]
$\lambda=1.0$ & 78.8 & 94.5 & 80.1 & 94.8\\ [.1em]
$\lambda=3.0$ & 76.6 & 93.2 & 78.4 & 94.1\\ [.1em]
\shline
\end{tabular}\vspace{2mm}
\caption{ Top-1 and top-5 linear classification accuracies (\%) on ImageNet-100 with different $\mathbf{\lambda}$'s.  The backbone network is ResNet-50.}
\label{table:labmda}\vspace{-8pt}
\end{table}
}

\def\tabTemp#1{
\begin{table}[#1]
\tablestyle{2.0pt}{1.0}
\setlength{\tabcolsep}{3pt}
\begin{tabular}{l||c|c|c|c|c|c}
\shline
$T (T_I = T_G)$ & 0.07 & 0.1 & 0.2 & 0.3 & 0.4 & 0.5 \\ [.1em]
\hline
CIFAR-100 & 57.9\% & 57.8\% & 58.1\% & 58.1\% & 57.6\% & 57.2\% \\
\hline
ImageNet-100 & 79.3\% & 79.6\% & 81.7\% & 80.7\% & 79.4\% & 79.0\%\\
\shline
\end{tabular}\vspace{2mm}
\caption{Linear (ImageNet-100) and kNN (CIFAR-100) evaluations for models trained with different choices of temperature $T$, $T_I = T_G$ for simplicity.}
\label{table:tau}\vspace{-8pt}
\end{table}
}

\def\tabVOC#1{
\begin{table}[#1]
\tablestyle{2.0pt}{1.0}
\begin{tabular}{l||cc|cc}
\shline
\multirow{2}{*}{pre-train methods} & \multicolumn{2}{c|}{VOC07+12} & \multicolumn{2}{c}{VOC07} \\
 & AP$_{50}$ & AP & AP$_{50}$ & AP \\
\hline
random initialization& 60.2 & 33.8  & - & - \\
supervised & 81.3 & 53.5 & 74.6 & 42.4 \\
\hline
JigSaw \cite{goyal2019scaling} & - & 53.3 & - & - \\
LocalAgg \cite{zhuang2019local}  & - & - & 69.1 & -\\
MoCo \cite{he2020momentum}  & 81.5 & 55.9 & 74.9 & 46.6\\
MoCo v2 \cite{chen2020improved} & 82.0 & 56.4 & - & - \\
\hline
NPID+CLD & 82.0 & 56.4 & 75.7 & 47.2 \\
MoCo+CLD & \bf{82.4} & \bf{56.7} & \bf{76.8} & \bf{48.3} \\
\shline
\end{tabular}\vspace{2mm}
\caption{Transfer learning results on object detection: We fine-tune on Pascal VOC $\textit{trainval07+12}$ or $\textit{trainval07}$, and test on VOC $\textit{test2007}$.  The detector is Faster R-CNN with ResNet50-C4.  MoCo v2 model is trained for 200 epochs with an MLP projection head.  Note that our model does not even use an  MLP projection head. Baseline results are from \cite{he2020momentum, chen2020improved}.}
\label{table:voc}
\end{table}
}

\renewcommand{\thefigure}{A.\arabic{figure}}
\renewcommand{\thetable}{A.\arabic{table}}

\section{Supplementary Materials}

We provide further details on 
Kitchen-HC construction, implementation details, and various choices and experiments we have explored to validate our approach.

\subsection{Kitchen-HC Dataset Construction} 
The original multi-view RGB-D kitchen dataset \cite{georgakis2016multiview} is comprised of densely sampled views of several kitchen counter-top scenes  with annotations in both 2D and 3D.  The viewpoints of the scenes are densely sampled and objects in the scenes are annotated with bounding boxes and in the 3D point cloud.  Kitchen-HC is constructed from multi-view RGB-D dataset Kitchen by extracting objects in their 2D bounding boxes. The customized Kitchen-HC dataset has 11 categories with highly correlated samples (from different viewing angles) and 20.8K / 4K / 14.4K instances for training / validation / testing.  \fig{kitchen} shows sample images in the original RGB-D Kitchen dataset from which our Kitchen-HC data are constructed (See samples used in Fig. 1).

\figKitchen{!h}

\subsection{Implementation Details}
We use SGD as our optimizer, with weight decay 0.0001 and momentum 0.9. We follow MoCo and NPID \cite{wu2018unsupervised,he2020momentum} and use only standard data augmentation methods for experiments on NPID+CLD and MoCo+CLD: random cropping, resizing, horizontal flipping, color and grayscale transformation, unless otherwise noticed.
\begin{enumerate}
\item {\bf ImageNet-\{100 \cite{tian2019contrastive}, ILSVRC-2012 \cite{deng2009imagenet} , Long-tail \cite{liu2019large}\}}.  For ILSVRC-2012 and ImageNet-LT, we use mini-batch size 256, initial learning rate 0.03, on 8 RTX 2080Ti GPUs. For ImageNet-100, we use batch size 512 and a larger initial learning rate of 0.8 on 8 GPUs, and apply the same setting to baselines and our methods. Training images are randomly cropped and resized to $224 \times 224$. For experiments on MoCov2+CLD with an MLP projection head, we extend the original augmentation in \cite{he2020momentum} by including the blur augmentation and apply cosine learning rate scheduler to further improve the performance on recognition as in \cite{chen2020improved}. BYOL+CLD is implemented based on OpenSelfSup \cite{openselfsup} benchmark.
For experiments on InfoMin+CLD and BYOL+CLD, we follow the same training recipe with InfoMin and BYOL \cite{tian2020makes,grill2020bootstrap} for fair comparisons.
\item {\bf CIFAR-\{10, 100, 10-LT, 100-LT\}, Kitchen-HC}. As \cite{wu2018unsupervised}, we use mini-batch size 256, initial learning rate  0.03 on 1 GPU for CIFAR \cite{krizhevsky2009learning} and Kitchen-HC. The number of epochs is 200 for CIFAR and 80 for Kitchen-HC.  Training images are randomly cropped and resized to $32 \times 32$.  
\item {\bf STL-10 \cite{coates2011analysis}}. Following \cite{ye2019unsupervised}, we use mini-batch size 256, initial learning rate 0.03, on 2 GPUs.  Baseline models and baselines with CLD are trained on "train+unlabelled" split (105k samples), and tested on "test" split (5k samples). Training images are randomly cropped and resized to $96 \times 96$.
\item {\bf Transfer learning on object detection}. We use Faster R-CNN with a backbone of R50-C4, with tuned synchronized batch normalization layers \cite{peng2018megdet} as the detector.  As in \cite{he2020momentum}, the detector is fine-tuned for 24k iterations for the experiment on Pascal VOC $\textit{trainval07+12}$ and 9k iterations for the experiment on Pascal VOC $\textit{trainval07}$. The image scale is [480, 800] pixels during training and 800 at inference. NPID+CLD and MoCo+CLD use the same hyper-parameters as in MoCo \cite{he2020momentum}. The VOC-style evaluation metric \cite{everingham2010pascal} AP$_{50}$ at IoU threshold is 50\% and COCO-style evaluation metric AP are used.
\item {\bf Semi-supervised learning}. To make fair comparisons with baseline methods, we use OpenSelfSup \cite{openselfsup} benchmark to implement baseline results and ours. We follow \cite{zhai2019s4l} and fine-tune the pre-trained model on two subsets for semi-supervised learning experiments, i.e. 1\% and 10\% of the labeled ImageNet-1k training datasets in a class-balanced way. The necks or heads are removed and only the backbone CNN is evaluated by appending a linear classification head. 

We apply greedy search on a list of hyper-parameter settings with the base learning rate from \{0.001, 0.01, 0.1\} and the learning rate multiplier for the head from \{1, 10, 100\}. We choose the optimal hyper-parameter setting for each method. Empirically, all baselines and their alternatives with CLD obtain the best performance with a learning rate of 0.01 and a learning rate multiplier for the head of 100.
We train the network for 20 epochs using SGD with weight decay 0.0001 and a momentum of 0.9, and a mini-batch of 256 on 4 GPUs. The learning rate is decayed by 5 times at epoch 12 and 16 respectively.
\end{enumerate}

\subsection{Which Clustering Method to Use?}
We have tried two popular clustering methods:  k-Means clustering and spectral clustering, both implemented in Pytorch for fast performance on GPUs. 

\noindent{\bf{k-Means clustering}} \cite{buchta2012spherical,kanungo2002efficient} aims to partition $n$ representations into $k$ groups, each representation belongs to the cluster with the nearest cluster centroid, serving as a prototype of the cluster.  We use spherical k-Means clustering which minimizes: $\sum(1 -\cos(f_i, u_{c(i)}))$
over all assignments $c$ of objects $i$ to cluster ids $c(i) \in \{1,...,k\}$ and over all prototypes $u_1,...,u_k$ in the same feature space as the feature vector $f_i$ representing the objects.  We use binary cluster assignment, where the cluster membership $m_{ij}=1$ if item $i$ is assigned to cluster $j$ and 0 otherwise.  The following k-means objective can be solved using the standard Expectation-Maximization algorithm \cite{dempster1977maximum}:
\begin{equation}
\begin{split}
    \Phi(M, \{u_1,...,u_C\}) &= \sum\limits_{i,j}m_{ij}(1 - \cos(f_i, u_{c(i)})) \\
    &= \sum\limits_{i,j}m_{ij}(1 - \frac{f_i\cdot u_{c(i)}}{||f_i||\cdot||u_{c(i)}||}).
\end{split}
\end{equation}

\noindent{\bf{Spectral clustering}} \cite{shi2000normalized, stella2003multiclass} treats data points as nodes of a graph.

\begin{enumerate}
    \item For feature $f_i \in \mathbb{R}^{d\times 1}$ of $N$ samples, we build a weighted gragh $\bf G = (V, E)$, with weight measuring pairwise feature similarity: $w_{i,j}=\frac{f_i\cdot f_j}{||f_i||\cdot ||f_j||}$.
    \item Let $\bf D$ be the $N\times N$ diagonal degree matrix with  $d_{ij} = \sum_{j=1}^n w_{ij}$ and  $\bf L$ be the normalized Laplacian matrix:
    \begin{equation}
    \mathbf{L = D^{-\frac{1}{2}}(D-W)D^{-\frac{1}{2}}}
    \end{equation}
    \item We compute the $k$ largest eigenvalues of $\mathbf{L}$ and use the corresponding $k$ row-normalized eigenvectors $E_i$ as the globalized new feature \cite{shi2000normalized, stella2003multiclass}.  We apply EM to find the cluster centroids. 
\end{enumerate}

Table \ref{table:cluster} shows that k-means clustering achieves better performance on Kitchen-HC and outperforms optimal spectral clustering result by 1.8\% when the group number is 10. However, as the group number increases, the performance difference becomes negligible.  
\tabClust{t}
\tabComponents{t}

\subsection{Are Separate Feature and Group Branches Necessary?} 
Intuitively, instance grouping and instance discrimination are at odds with each other.  Our solution is to formulate the feature learning on a common representation, forking off two branches where we can impose grouping and discrimination separately.
Table \ref{table:components} shows that projecting the representation to different spaces and jointly optimize the two losses increase top-1 kNN accuracy by 1.5\% and 1.8\% on CIFAR-10 and CIFAR-100 respectively.

\subsection{How Effective Is Cross-Augmentation Comparisons?}

Instance-level discrimination presumes each instance is its own class and
any other instance is a negative.  The groups needed for any group-level discrimination have to be built upon local clustering results extracted from the current feature in training, which are fluid and unreliable. 

Our solution is to seek the most certainty among all the uncertainties:
We presume stable grouping between one instance and its augmented version, and our cross-level discrimination compares the former with the groups derived from the latter.  We roll the three processes: instance grouping, invariant mapping, and instance-group discrimination all into one CLD loss.

Table \ref{table:components} shows that our cross-augmentation comparison increases the top-1 accuracy by more than 2\% on recognition task.  It demands the feature not only to be invariant to data augmentation, but also to be respectful of natural grouping between individual instances, often aligning better with downstream semantic classification.

\subsection{How Sensitive Are Hyper-parameters Weight $\mathbf{\lambda}$ and Temperature $T$?} 
$\lambda$ controls the relative importance of CLD with respect to instance-level discrimination, and helps strike a balance between the caveates of noisy initial grouping and the benefits it brings with coarse-grained repulsion between instances and local groups.  
Table \ref{table:labmda} shows that, at a fixed group number,  $\lambda=0.25$ achieves optimal performance, and a larger $\lambda$ generally leads to worse performance and even decreases top-1 accuracy by 3.1\% at $\mathbf{\lambda}=3$. 

\tabLambda{h}

$T$ is known to critical for discrminative learning and can be sometimes tricky to choose.  Table \ref{table:tau} shows that the best performance is achieved at $T=0.2$ for both CIFAR and ImageNet-100.
 With local grouping built into our CLD method, we find the sensitivity of $T$ is greatly reduced.

\tabTemp{h}

\subsection{Is A Larger Memory Bank Always Better for Discriminative Learning?}
%

A larger memory bank includes more negatives and is known to deliver a better discriminator.  However, we cannot simply adjust the memory bank size according to NMI or retrieval accuracy in order to deliver the best performance on downstream classification.

\fig{infoNCEK} compares NMI and retrieval accuracies  under different negative prototype numbers.  If there are too many negatives, the model would focus on repelling negative instances, ignoring the commonality between instances; if there are too few negatives, the model would be subject to random fluctuations from batch to batch, affecting optimization and convergence. 
However, neither the number of negatives (i.e. infoNCE-k) to obtain the best retrieval accuracy nor the number of negatives to achieve the best NMI score can deliver the best downstream classification task. To deliver optimal performance at downstream classification task, there is a trade-off between local mutual information (evaluated by retrieval task) and global mutual information (evaluated by Normalized Mutual Information).

\figinfoNCEK{!t}

\figRetrieval{h}

\subsection{Sample Retrievals}
\fig{Retrieval} shows our near-perfect sample retrievals on ImageNet-100 using $f_I(x)$ in our NPID + CLD model. On the contrary, NPID seems to be much more sensitive to textural appearance (e.g., Rows 1,4,6,7), first retrieve those with similar textures or colors.  CLD is able to retrieve semantically similar samples.  Our conjecture is that by gathering similar textures into groups, CLD can actually find more informative feature that contrasts between groups. For example, the 5th query image is a Chocolate sauce, which has similar texture with Grouper fish. NPID incorrectly retrieves many images from the Grouper Fish class, but CLD successfully captures the semantic information of the query image, and retrieves instances with the same semantic information.

\end{document}